\documentclass{article}
\usepackage{iclr2027_conference,times}
\usepackage{amsmath,amsfonts,bm}

\def\eqref#1{equation~\ref{#1}}
\def\1{\bm{1}}

\DeclareMathAlphabet{\mathsfit}{\encodingdefault}{\sfdefault}{m}{sl}
\SetMathAlphabet{\mathsfit}{bold}{\encodingdefault}{\sfdefault}{bx}{n}

\usepackage{amsmath,amssymb,mathtools}
\usepackage{booktabs}
\usepackage{multirow}
\usepackage{array}
\usepackage{xcolor}
\usepackage{microtype}
\usepackage{hyperref}
\usepackage{url}
\usepackage{amsthm}
\usepackage{amsmath}
\usepackage{amssymb}
\usepackage{wrapfig}
\usepackage{subcaption}
\usepackage{tabularx}
\usepackage{array}
\usepackage{longtable}
\usepackage{graphicx}
\theoremstyle{definition}
\newtheorem{definition}{Definition}
\newtheorem{problem}{Problem}

\usepackage{chngcntr}
\usepackage{fancyhdr}
\usepackage{placeins}

\fancypagestyle{arxivstyle}{
  \fancyhf{}
  \renewcommand{\headrulewidth}{0pt}
  \renewcommand{\footrulewidth}{0pt}
  \cfoot{\thepage}
}

\hypersetup{hidelinks}

\hypersetup{
  pdftitle={HazardWeaver: Scientific Route Selection for Hazard Analysis Agents},
  pdfauthor={Wangshu Zhu, Xueqi Cheng, Liang Wu, Yushun Dong}
}
\vspace{4em}
\newcommand{\makearxivtitle}{
  \begin{center}
    \rule{\textwidth}{3pt}

    {\LARGE\bfseries
    HazardWeaver: Scientific Route Selection\\
    for Hazard Analysis Agents\par}

    \rule{\textwidth}{0.8pt}

    {\large
    Wangshu Zhu$^{1}$ \quad
    Xueqi Cheng$^{1}$ \quad
    Liang Wu$^{2}$ \quad
    Yushun Dong$^{1}$\par}

    $^{1}$Department of Computer Science, Florida State University
    \qquad
    $^{2}$Nokia

    {\ttfamily
    \{wz26b,xc25,yushun.dong\}@fsu.edu
    \qquad
    liang.wu@nokia.com\par}
  \end{center}

}
\iclrfinalcopy

\begin{document}

\pagestyle{arxivstyle}
\thispagestyle{arxivstyle}

\makearxivtitle
\raggedbottom

\begin{abstract}
Understanding and assessing natural hazards is essential for disaster preparedness and risk reduction. Recent advances in large language models have spurred growing interest in AI agents for hazard analysis, particularly their ability to integrate scientific data, models, and tools into automated workflows. However, effective automation requires agents to determine which scientific methods are appropriate for a given event and executable with the available data and tools. As new evidence and execution results become available, these conditions can change, requiring agents to reconsider their choices. We formulate this problem as state-dependent scientific route selection and introduce HazardWeaver. Specifically, HazardWeaver first leverages the Hazard Knowledge Compiler to extract evidence-linked conditions governing scientific applicability, then its Hazard Capability Graph represents executable scientific capabilities and checks compatibility between their inputs and outputs. Using these complementary representations, the Hazard Weaver Agent component selects applicable and executable routes, carries out their workflows, and revises its decisions as the analysis state changes. To evaluate both the scientific outputs and the decisions that produce them, we introduce the Hazard Weaver Benchmark, comprising 141 instances across seven single-hazard domains and four multi-hazard interaction classes. 
The benchmark accommodates multiple valid scientific routes and evaluates output correctness, route validity, and justified abstention. Extensive experiments on this benchmark show that HazardWeaver outperforms existing agent systems, with the largest gains on tasks with multiple eligible scientific routes. Our code is publicly available at \url{https://github.com/LabRAI/HazardWeaver}.
\end{abstract}
\section{Introduction}
\label{sec:introduction}

Natural-hazard analysis supports disaster preparedness and risk reduction. The same target can often be reached through several scientific routes. Wildfire spread, flood extent, and earthquake source parameters, for example, can each be estimated through empirical, physical, statistical, or learned methods, which differ in assumptions, required observations, validity ranges, and computational dependencies \citep{perry_current_1998,huot_next_2022,nearing_global_2024,zhu_phasenet_2018,stewart_selection_2015,reichstein_deep_2019,karpatne_theoryguided_2017}. A task specification alone therefore does not fix how it should be analyzed, since scientific evidence determines which routes apply while the available resources determine which routes can currently run. The difficulty grows when multiple hazards occur simultaneously or in cascade, since one hazard alters the conditions under which another must be assessed \citep{kappes_challenges_2012,gill_reviewing_2014a,zscheischler_future_2018,zscheischler_typology_2020}.
Recent advances in large language models have spurred growing interest in AI agents that integrate such scientific data, models, and tools into automated workflows \citep{boiko_autonomous_2023a,m.bran_augmenting_2024a,jaber2026autoclimds}. On an actual event, however, the evidence and tools an agent depends on can shift as the event unfolds. The agent must therefore repeatedly reconsider which route it can trust and run. Route choice thus becomes a state-dependent decision, which leads to the question we study: given a hazard task and the current analysis state, which scientific route should an agent take?

% On an actual event, however, the evidence and tools available to an agent can change as the analysis unfolds. The agent must therefore repeatedly reassess which routes remain scientifically applicable and operationally executable.

Existing work provides important components for automated hazard analysis, but two gaps remain in how scientific routes are grounded and revised. \textit{Gap 1: Scientific Route Grounding.} Hazard models offer established procedures within selected modeling regimes \citep{finney_farsite_1998,bates_simple_2010}, yet the conditions governing their applicability are scattered across papers, reports, and model documentation. Selecting an appropriate route therefore requires making these conditions explicit and checking them against the available task evidence. This grounding must also extend to how scientific assets are combined. Although workflow and disaster-agent systems expose executable tools and operational pipelines, a route may draw on heterogeneous assets whose outputs and inputs are not directly compatible. These assets must therefore be represented as executable capabilities with explicit applicability conditions and compatibility requirements. \textit{Gap 2: State-Dependent Route Selection.} Automated experimentation methods, from classical AutoML search \citep{akiba_optuna_2019,wang_flaml_2021} to recent LLM-driven data-science agents, optimize models, programs, or workflows for a specified objective. However, this optimization alone does not establish which routes remain scientifically and operationally eligible as execution proceeds. In hazard analysis, new artifacts, unavailable inputs, execution failures, or changing downstream requirements can alter the eligible route set. An agent must therefore reassess route eligibility as the scientific and computational state evolves and revise its decisions accordingly. These requirements also call for route-aware evaluation: because several routes may validly satisfy the same task, evaluation must assess both final-artifact quality and whether route selection, revision, or abstention was justified under the observed state.

\begin{wrapfigure}{r}{0.47\columnwidth}
  \vspace{-2.5\baselineskip}
  \centering
  \includegraphics[width=\linewidth]{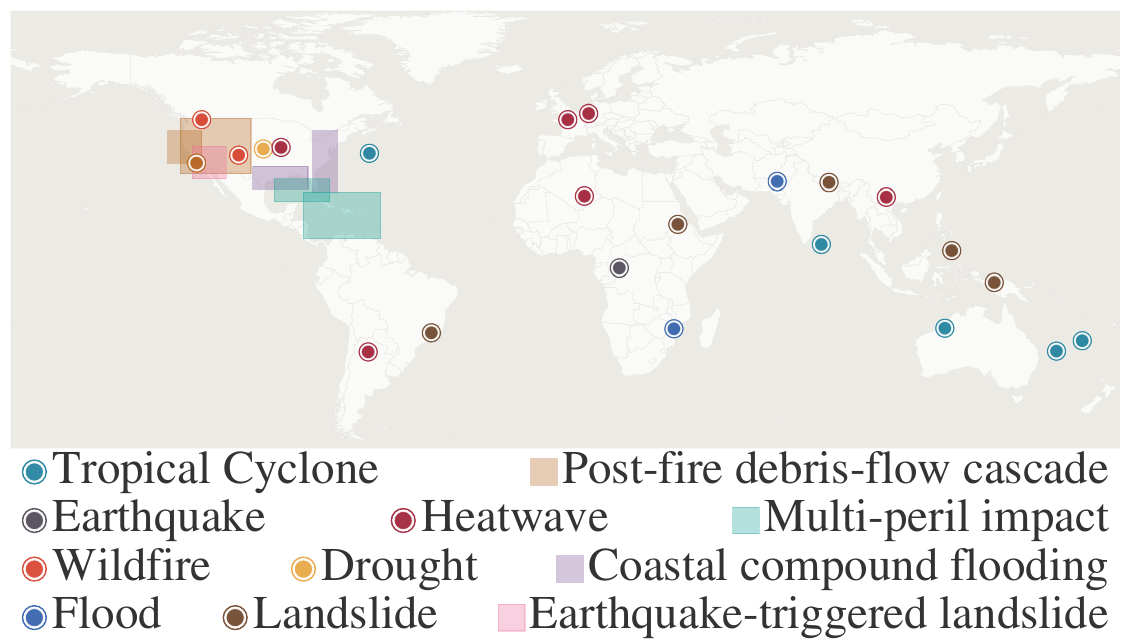}
  \vspace{-1.6\baselineskip}
  \caption{Global coverage of HWB.}
  \label{fig:hwb-world-map}
  \vspace{-1.4\baselineskip}
\end{wrapfigure}
To meet these requirements, \textsc{HazardWeaver} integrates scientific grounding, state-dependent route selection, and route-aware evaluation within a unified evidence-to-execution architecture. Scientific route grounding is supported by two complementary components. The Hazard Knowledge Compiler (HKC) derives evidence-linked applicability conditions from scientific literature and resolves them against the task context. The Hazard Capability Graph (HCG) represents hazard models, datasets, formulas, and transformations as typed executable capabilities, connecting them only under a registered compatibility relation. Building on these components, the Hazard Weaver Agent (HWA) determines the eligible route set by combining the applicability conditions from HKC with capability reachability in HCG. It selects and executes a route, then reassesses eligibility and revises its decisions as the state evolves. Within this process, the language model manages route decisions and execution flow, while scientific models and solvers produce the numerical, spatial, and probabilistic outputs. To evaluate both the resulting artifacts and the decisions that produce them, the Hazard Weaver Benchmark (HWB) assesses artifact correctness, route validity at commitment time, and recovery-aware abstention. Specifically, HWB contains 141 sealed instances spanning seven single-hazard domains and four multi-hazard interaction classes. Its construction infrastructure includes a global atlas of 3,062 registered geospatial scenarios, with 542 materialized for benchmark generation (Figure~\ref{fig:hwb-world-map}).

\paragraph{Contributions.}
Our contributions can be summarized as follows:
\begin{itemize}
\item \textbf{A novel scientific-agent decision problem.} We formulate hazard analysis as state-dependent scientific route selection, where route eligibility requires both evidence-based scientific applicability and capability reachability, and is updated throughout execution.

\item \textbf{An evidence-grounded adaptive agent system.} We introduce an agent system that compiles evidence into route conditions, connects scientific capabilities through typed compatibility, and updates executable workflows as the analysis state evolves.

\item \textbf{A globally grounded, route-aware benchmark.} HWB comprises 141 instances built from scientific datasets and benchmarks, spans seven single-hazard domains and four multi-hazard interaction classes, and jointly evaluates artifact correctness and route validity.
\end{itemize}

\section{Preliminaries}
\label{sec:problem}
Let $\mathcal{H}$ denote the set of hazard settings, each being either a single hazard or a multi-hazard interaction class. For $h\in\mathcal{H}$, let $\mathcal{T}_h$ denote its associated scientific tasks and $\mathcal{R}_t$ the registered route families for task $t\in\mathcal{T}_h$, together forming the design space $\mathcal{M}=\{(h,t,r): h\in\mathcal{H},\,t\in\mathcal{T}_h,\,r\in\mathcal{R}_t\}$. A task instance is $q=(h,t,\xi,\Gamma_t,c_q)$, where $\xi$ is the event and environmental context, $\Gamma_t$ specifies the requested output in $\mathcal{Y}_t$, and $c_q$ is the information cutoff associated with the decision. We denote scientific knowledge by $\mathcal{K}$ and executable scientific capabilities by $\mathcal{G}$.

% Let $\mathcal{H}$ denote the set of hazards, $\mathcal{T}_h$ the scientific tasks associated with hazard $h\in\mathcal{H}$ \xq{h here looks like one hazard, maybe $h_i$ or something else?}, and $\mathcal{R}_t$ the registered route families for task $t\in\mathcal{T}_h$, together forming the design space $\mathcal{M}=\{(h,t,r): h\in\mathcal{H},\,t\in\mathcal{T}_h,\,r\in\mathcal{R}_t\}$. A task instance is $q=(h,t,\xi,\Gamma_t,c_q)$, where $\xi$ is the event and environmental context, $\Gamma_t$ specifies the requested output in $\mathcal{Y}_t$, and $c_q$ is the information cutoff associated with the decision. We denote scientific knowledge by $\mathcal{K}$ and executable scientific capabilities by $\mathcal{G}$.

\begin{definition}[Route family and workflow]
\label{def:route}
A route family $r \in \mathcal{R}_t$ specifies a scientific solution structure together with its scientific conditions and required capability signatures. A concrete workflow $\pi \in \operatorname{Inst}(r;\mathcal{G})$ realizes $r$ by binding its stages to registered datasets, models, formulas, transformations, software, and other capabilities in $\mathcal{G}$.
\end{definition}

\begin{definition}[Analysis state and trajectory]
\label{def:state}
At decision step $k$, the analysis state is $s_k=(\mathcal{X}_k,\mathcal{C}_k,b_k)$, where $\mathcal{X}_k$ contains available raw and derived artifacts, $\mathcal{C}_k$ records capability availability, and $b_k$ is the remaining resource budget. An execution trajectory is $\tau=(s_0,a_0,s_1,\ldots,a_{K-1},s_K)$, and route commitments made along $\tau$ are recorded as $\operatorname{Commit}(\tau)=\{(k,r_k)\}$.
\end{definition}

A route counts as usable at $s_k$ only when it clears two checks: scientific applicability $A_{\mathrm{sci}}(r,q,s_k;\mathcal{K})$, resolved from the evidence attached to $r$, and capability reachability $A_{\mathrm{cap}}(r,s_k;\mathcal{G})$, resolved from the executable assets available at $s_k$; Section~\ref{sec:method} details how each operator is computed. The eligible route set at $s_k$ collects exactly the routes passing both checks,
\begin{equation}
\mathcal{R}_t^{\mathrm{elig}}(q,s_k) \;=\; \bigl\{\, r\in\mathcal{R}_t \,:\, A_{\mathrm{sci}}(r,q,s_k;\mathcal{K})=\textsc{Applicable} \ \wedge\ A_{\mathrm{cap}}(r,s_k;\mathcal{G})=\textsc{Reachable} \,\bigr\}.
\label{eq:eligible}
\end{equation}

\begin{problem}[State-Dependent Scientific Route Selection and Execution]
\label{prob:main}
Given $(q,s_0,\mathcal{K},\mathcal{G},B)$, the goal at every committed step $k$ is to execute a route $r_k$ drawn from $\mathcal{R}_t^{\mathrm{elig}}(q,s_k)$ (Eq.~\ref{eq:eligible}), recomputing this set as $s_k$ evolves with new evidence, artifacts, and capability availability, and returning $y\in\mathcal{Y}_t$ or abstaining with $\bot$ once no state reachable within $B$ still admits one.
\end{problem}

\section{Methodology}
\label{sec:method}

\begin{figure*}[t]
    \centering
    \includegraphics[width=\textwidth]{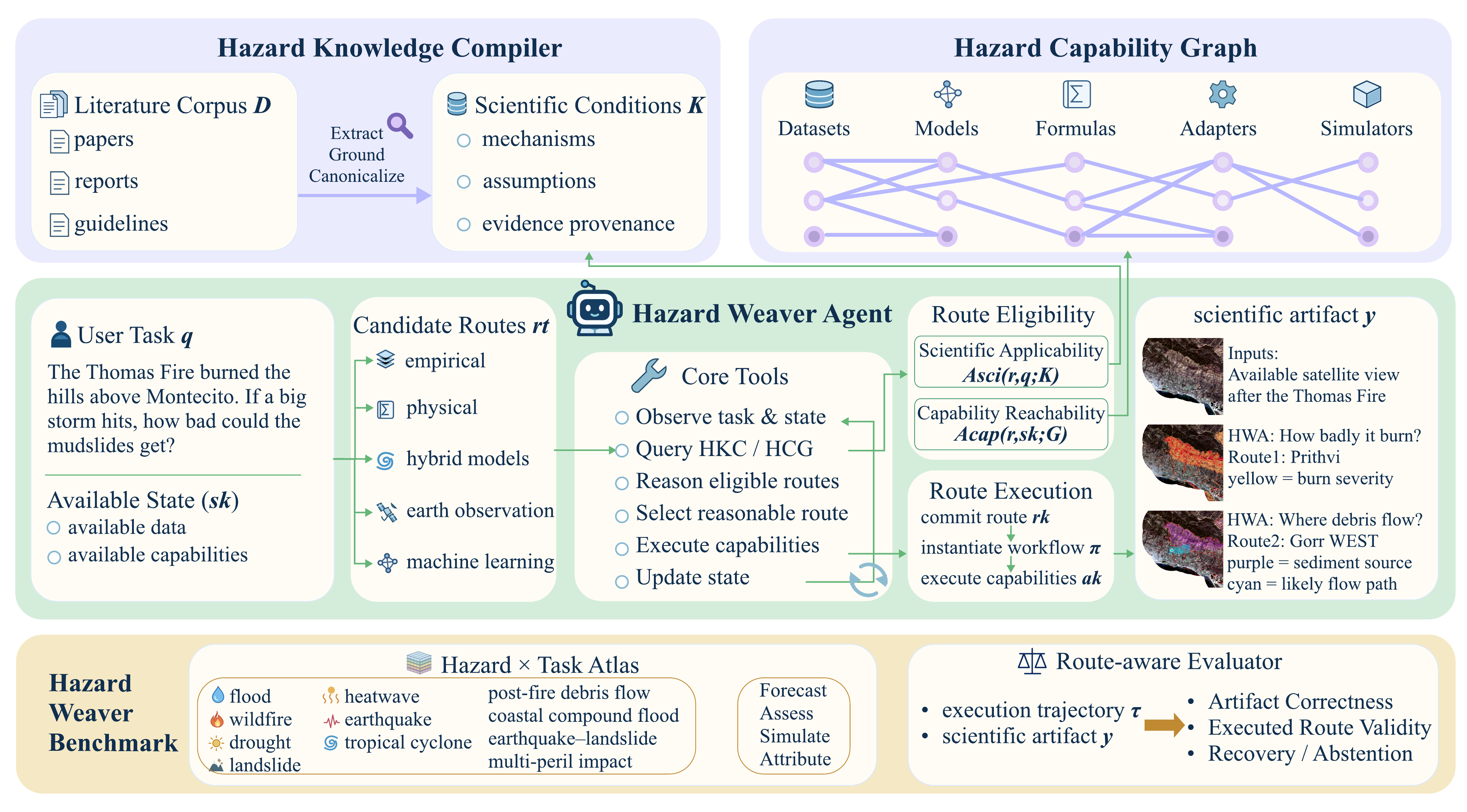}
    \vspace{-0.75cm}
    \caption{Overview of HazardWeaver.
    HKC and HCG identify scientifically applicable and currently executable routes, whose intersection defines the eligible route set for HWA.
    HWA selects, executes, and revises model-backed routes as the analysis state changes, while HWB independently evaluates the resulting artifact, route validity, and recovery or abstention.}
    \label{fig:hw_framework}
    \vspace{-0.4cm}
\end{figure*}

This section develops \textsc{HazardWeaver}, shown in Figure~\ref{fig:hw_framework}, in the order its route decision actually forms. Section~\ref{sec:hkc} grounds scientific applicability through HKC, Section~\ref{sec:hcg} establishes capability reachability through HCG, Section~\ref{sec:weaver-agent} combines both signals in HWA to select, execute, and revise a route, and Section~\ref{sec:hwb} evaluates the resulting decisions through HWB.

% \xq{there are still too many texts in the overview plot, let's use some icons to make this easier to follow }

\subsection{Hazard Knowledge Compiler: Scientific Applicability}
\label{sec:hkc}
\looseness=-1
A scientific route applies only when its assumptions and validity conditions match the task context, evidence that sits scattered across papers and model documentation. HKC extracts a structured condition set from its current literature corpus and checks it against the available task evidence. Retrieval-augmented methods provide relevant context \citep{lewis_retrievalaugmented_2020}, while scientific information extraction provides representations of entities and relations \citep{luan_multitask_2018,jain_scirex_2020}.

\textbf{Evidence-Grounded Condition Extraction.} For each route family $r$, HKC retrieves this evidence, maps it into a common schema, and canonicalizes equivalent entities into a condition set $\mathcal{K}_r=\{\kappa_1,\ldots,\kappa_{m_r}\}$, where each $\kappa_i$ records a proposition's semantic role, applicable scope, logical direction, and supporting source, spanning route assumptions, required observations, validity ranges, scale, uncertainty, and known failure conditions. 
\textbf{Condition-Level Applicability.}  At task time, HKC relates each required condition $\kappa_i\in\mathcal{K}^{\mathrm{req}}_r$ to the evidence available in $\xi$ and $s_k$, assigns $z_i(q,s_k)\in{\textsc{Satisfied},\textsc{Violated},\textsc{Unresolved}}$, and derives route-level applicability:
\begin{equation}
A_{\mathrm{sci}}(r,q,s_k;\mathcal{K})
=
\begin{cases}
\textsc{Applicable},
& z_i(q,s_k)=\textsc{Satisfied},
  \ \forall \kappa_i\in\mathcal{K}^{\mathrm{req}}_r,\\
\textsc{Inapplicable},
& \exists\,\kappa_i\in\mathcal{K}^{\mathrm{req}}_r:
  z_i(q,s_k)=\textsc{Violated},\\
\textsc{Undetermined},
& \text{otherwise}.
\end{cases}
\label{eq:sci-app}
\end{equation}
The task target and information cutoff remain fixed; condition status is recomputed from evidence available in the current state. Applicability is thus grounded in individual condition status, each carrying its own provenance and task evidence, so an unresolved or changed condition can revise the route decision without reconstructing the rest of the record; an \textsc{Undetermined} condition keeps $r$ outside the eligible set (Eq.~\ref{eq:eligible}) without discarding it, since HWA can request the missing evidence during recovery (Section~\ref{sec:weaver-agent}) and re-resolve $A_{\mathrm{sci}}$ once new context arrives.

\subsection{Hazard Capability Graph: Scientific Capability Composition}
\label{sec:hcg}

Scientific applicability does not guarantee executability, since an appropriate route can still require unavailable capabilities or produce artifacts that no downstream model can consume. Analysis-readiness specifications motivate explicit scientific metadata \citep{baumann_analysisreadiness_2024}, and semantic workflow systems support composition under component constraints \citep{gil_wings_2011}. HCG represents the available scientific ecosystem as a typed directed graph $\mathcal{G}=(\mathcal{V}_{G},\mathcal{E}_{G})$ and checks whether a route's required capabilities are actually reachable in the current state.

\textbf{Typed Capability Composition.} Nodes in $\mathcal{G}$ are executable capabilities, each recording its entry point, scientific function, dependencies, validation metadata, provenance, and typed input and output ports of type $\alpha=(v,u,f,\Omega_x,\Omega_t,\delta,\sigma,p)$, and edges are valid artifact transfers: for a produced artifact of type $\alpha'$ and a required input type $\alpha$, a direct connection is permitted only when
\begin{equation}
\alpha'\models\alpha
\quad\Longleftrightarrow\quad
{\textstyle\bigwedge_{j=1}^{8}}\alpha'_j\sim_j\alpha_j.
\label{eq:type-compat}
\end{equation}
where $\sim_j$ is the registered compatibility relation for field $j$; exact agreement and representation-preserving conversions satisfy this directly, while information-changing operations such as aggregation, interpolation, learned downscaling, and thresholding remain explicit graph capabilities, so their effects stay visible in the workflow's provenance. \textbf{State-Dependent Capability Reachability.} Given route family $r$ and state $s_k$, HCG searches for workflow realizations $\operatorname{Inst}(r;\mathcal{G},s_k)$ whose required capabilities are available and whose adjacent artifacts satisfy Eq.~\ref{eq:type-compat}, giving capability reachability
\begin{equation}
A_{\mathrm{cap}}(r,s_k;\mathcal{G})
=
\begin{cases}
\textsc{Reachable},
& \operatorname{Inst}(r;\mathcal{G},s_k)\neq\varnothing,\\
\textsc{Unavailable},
& \operatorname{Inst}(r;\mathcal{G},s_k)=\varnothing
  \text{ under the known state},\\
\textsc{Undetermined},
& \text{capability information is incomplete}.
\end{cases}
\label{eq:cap-reach}
\end{equation}
As with $A_{\mathrm{sci}}$, an \textsc{Undetermined} status keeps $r$ outside the eligible set without discarding it. When capability availability or an artifact changes, HCG searches for a compatible replacement subpath or prompts HWA to reconsider the route when it does not.

Provenance records connect each derived artifact to the capabilities and inputs that produced it, following the purpose of interoperable workflow provenance \citep{khan_sharing_2019}. The illustrated scientific assets include an Earth-observation foundation model \citep{jakubik_foundation_2023} and WEST, a postfire debris-flow volume model \citep{gorr_improved_2026}. A spatial flow-path artifact requires a separate downstream capability in addition to a volume estimate.

\subsection{Hazard Weaver Agent: Adaptive Route Selection}
\label{sec:weaver-agent}

HWA operationalizes Eq.~\ref{eq:eligible} at every decision step, turning the intersection of scientific applicability and capability reachability into an executed, revisable route commitment.

\textbf{Model-Backed Route Execution.} At state $s_k$, $\mathcal{R}_t^{\mathrm{elig}}(q,s_k)$ combines $A_{\mathrm{sci}}$ from HKC with $A_{\mathrm{cap}}$ from HCG, which exclude routes for different reasons, conflicting scientific requirements versus requirements that cannot currently be realized, letting HWA trace an excluded route to its specific cause and target the matching recovery operation. For every $r$ in this set, HWA receives its grounded conditions, workflow realizations, provenance, uncertainty, and resource requirements, and a language-model policy selects $r_k\sim\mu_\theta(\cdot\mid q,s_k,\mathcal{R}_t^{\mathrm{elig}}(q,s_k))$, records the commitment $(k,r_k)$, instantiates $\pi^{(k)}\in\operatorname{Inst}(r_k;\mathcal{G},s_k)$, and executes its capabilities; the language model controls only the route decision and execution flow, while the executed models, solvers, formulas, and transformations produce the numerical, spatial, and probabilistic outputs, keeping the terminal artifact linked through provenance to the route that produced it. \textbf{Revisable Route Commitments.} Execution can reveal missing inputs, alter capability availability, or invalidate downstream compatibility, so HWA updates $s_k$ and recomputes Eq.~\ref{eq:eligible} after every step: a route that remains eligible continues with its current or repaired workflow, and one that does not sends the policy back to the updated eligible set. When this set is empty, HWA explores permitted recovery operations, information requests, compatible transformations, capability substitutions, and alternative realizations, over the states $\operatorname{Rec}(s_k,b_k)$ reachable within the remaining budget, and uses the recovery outcome to decide whether to abstain. Here $\operatorname{Rec}(s_k,b_k)$ is the budget-bounded reachable closure over the finite set of grounded actions permitted for the instance. The implemented search explores this closure within its budget. Failure to find a route is not a formal proof that no feasible route exists.

\subsection{Hazard Weaver Benchmark: Route-Aware Evaluation}
\label{sec:hwb}

A task may admit several valid routes, and the valid set may shift as conditions or capabilities change, so artifact correctness alone cannot certify a valid decision. We therefore introduce the \textbf{Hazard Weaver Benchmark} (HWB), which evaluates the final artifact together with the route and execution state that produced it. HWB derives tasks with explicit output contracts from established scientific datasets, operational products, and benchmarks, and constructs candidate routes from evidence-grounded Route Cards and registered capabilities. Instances are admitted only after provenance, temporal-cutoff, and replay checks, after which the inventory, route registry, and task-specific evaluators are sealed; benchmark-side references and route judgments remain hidden from the solver. Appendix~\ref{app:hwb-scope} and Appendix~\ref{app:evaluation-protocol} detail the construction and evaluation protocol. 

% A task may admit several valid routes, and the valid set may shift as conditions or capabilities change, so artifact correctness alone cannot certify a valid decision. \xq{Therefore, in this paper we formally introduce \textbf{Hazard Weaver Benchmark} to help stakeholders assess} the final artifact together with the route and execution state that produced it. \xq{Below we may also need to talk about how we collect the data, set up the evaluation metric, etc briefly, and say the details are in appendix xxx? }

\textbf{Benchmark-Side Route Validity.} Using hidden scientific conditions and benchmark-side capability state kept outside solver context, HWB constructs $\mathcal{R}_t^{\mathrm{elig},B}(q,s_k)$ by the same criteria as Eqs.~\ref{eq:sci-app}--\ref{eq:cap-reach}. A trajectory is route-valid when every commitment falls inside this set at its commitment state,
\begin{equation}
V_{\mathrm{route}}^{B}(q,\tau)
=
\mathbb{1}
\left[
r_k\in\mathcal{R}_t^{\mathrm{elig},B}(q,s_k),
\;
\forall(k,r_k)\in\operatorname{Commit}(\tau)
\right],
\label{eq:route-valid}
\end{equation}
 crediting any route the benchmark itself registers as valid without requiring the agent to reproduce the construction-time witness workflow. Writing $\theta(x)$ for a raw artifact's information time and $\operatorname{Src}(z)$ for the raw-source ancestors of derived artifact $z$, temporal validity requires every such ancestor to have been available by the cutoff $c_q$,
 \begin{equation}
 V_{\mathrm{causal}}(q,\tau)
=\mathbb{1}\Big[\theta(x)\le c_q,\ \forall x\in{\textstyle\bigcup_{z\in\operatorname{Used}(\tau)}}\operatorname{Src}(z)\Big],
 \label{eq:causal}
 \end{equation}
 so a derived artifact may be computed after $c_q$ once all its sources were available in time. 
 Together with $V_{\mathrm{type}}(\tau)$ from Eq.~\ref{eq:type-compat}, provenance validity $V_{\mathrm{prov}}(y,\tau)$ linking the terminal artifact to an executed route commitment, and budget validity $V_{\mathrm{budget}}(\tau;B)$, these give joint execution validity $V_{\mathrm{exec}}(q,\tau;B)=V_{\mathrm{type}}(\tau)\cdot V_{\mathrm{causal}}(q,\tau)\cdot V_{\mathrm{budget}}(\tau;B)$. The benchmark contract accepts abstention only when permitted recovery cannot reach another registered benchmark-valid route,
 \begin{equation}
 V_{\mathrm{abs}}^{B}(q,s_k,b_k)
=\mathbb{1}\Big[{\textstyle\bigcup_{s'\in\operatorname{Rec}(s_k,b_k)}}\mathcal{R}_t^{\mathrm{elig},B}(q,s')=\varnothing\Big],
 \label{eq:abstention}
 \end{equation}
 which defines the benchmark criterion over its registered recovery transitions. Applying this criterion requires the benchmark to establish the reachable set under those transitions; a solver timeout alone does not establish that no usable route remains.

\textbf{Decision-Constrained Accuracy.} HWB combines artifact correctness with these validity checks into Decision-Constrained Accuracy: writing the solver's return as $o\in\mathcal{Y}_t\cup\{\bot\}$ and $E_q(y)\in\{0,1\}$ for task-specific artifact success under its native evaluator,
\begin{equation}
\mathrm{DCA}(q,o,\tau;B)
=
\begin{cases}
E_q(y)\,
V_{\mathrm{route}}^{B}(q,\tau)\,
V_{\mathrm{prov}}(y,\tau)\,
V_{\mathrm{exec}}(q,\tau;B),
& o=y,\\[3pt]
V_{\mathrm{abs}}^{B}(q,s_K,b_K)\,
V_{\mathrm{route}}^{B}(q,\tau)\,
V_{\mathrm{exec}}(q,\tau;B),
& o=\bot,
\end{cases}
\label{eq:dca}
\end{equation}
aggregated over an evaluation set as $\mathrm{DCA}(\mathcal{Q})=\frac{1}{|\mathcal{Q}|}\sum_{q\in\mathcal{Q}}\mathrm{DCA}(q,o_q,\tau_q;B)$; native scientific metrics stay task-specific and are reported separately, so DCA measures decision validity without folding heterogeneous scientific errors into one score. HWB further builds matched interventions that alter one factor governing route choice while holding the rest of the task fixed: evidence relevant to $A_{\mathrm{sci}}$, availability or compatibility relevant to $A_{\mathrm{cap}}$, the state after execution begins, or an upstream hazard variable in a registered physical interaction. These interventions test whether route selection, recovery, and the resulting artifacts respond consistently to controlled changes in the route space.

\section{Experimental Setup}
\label{sec:experimental-setup}

\textbf{Benchmark.}
The primary comparison uses the same 141 sealed HWB instances, HCG capability catalog, and evaluator for all systems. The instances draw on eleven published datasets, operational products, and scientific benchmarks: PWFDF and USGS Ground Failure \citep{king_post-wildfire_2025,allstadt_us_2022}, SFINCS and FloodCastBench \citep{leijnse_modeling_2021,xu_floodcastbench_2025}, NOAA CPC SPI \citep{noaa_cpc_spi_2026}, SeisBench \citep{woollam_seisbench_2022}, ExtremeWeatherBench and NASA LHASA v2 \citep{mcgovern_extreme_2026,nasa_global_landslide_nowcast_2025}, TCBench and NOAA Storm Events \citep{gomez_tcbench_2026a,noaa_ncei_storm_events_1996}, and WildfireSpreadTS \citep{gerard_wildfirespreadts_2023}. These sources provide observed events, forcing data, and reference outputs used to construct route-selection and execution tasks. The inventory contains 90 single-hazard and 51 multi-hazard instances; 45 admit one route and 96 admit multiple routes. Appendix~\ref{app:hwb-scope} defines the tracks and evaluation strata.

\textbf{Baselines.}
We compare HazardWeaver with adapted versions of Plan-and-Solve \citep{wang_planandsolve_2023}, Self-Consistency \citep{wang_selfconsistency_2023}, DisasterBench-ToT \citep{chen_disasterbench_2026,yao_tree_2023}, MLE-STAR \citep{nam_mlestar_2025}, AutoML-Agent \citep{trirat_automlagent_2025}, Reflexion \citep{shinn_reflexion_2023}, AIDE \citep{jiang_aide_2025}, DS-Agent \citep{guo_dsagent_2024}, R\&D-Agent \citep{yang_rdagent_2025}, and ReAct \citep{yao_react_2022}. All methods use the same task inventory, policy backbone, route catalog, scientific replay, and evaluator. Appendix~\ref{app:baseline-details} documents the adaptations and implementation details.

\textbf{Metrics.}
The primary metric is Decision-Constrained Accuracy (DCA; Eq.~\ref{eq:dca}). We also report artifact validity $E_q$, aggregate trajectory validity $V_q$, and Macro DCA, which weights the 11 tracks equally. Binomial rates use Wilson 95\% confidence intervals \citep{wilson_probable_1927}.

\textbf{Implementation.}
The five ablation backbones are OLMo-3.1-32B, Mixtral-8x22B, Gemma-4-31B-it, Llama-3.3-70B, and DeepSeek-V4.1-Flash. The corresponding model-family reports describe OLMo \citep{olmo_olmo_2026}, Mixtral \citep{jiang_mixtral_2024}, Gemma \citep{gemma_gemma_2026}, Llama \citep{grattafiori_llama_2024}, and DeepSeek \citep{deepseek_deepseekv41flash_2026}. The evaluated variants are listed in Appendix~\ref{app:implementation-details}.

\section{Experiments}
\label{sec:experiments}

We evaluate \textsc{HazardWeaver} through three research questions.
\textbf{RQ1:} How does HazardWeaver compare with representative agent baselines?
\textbf{RQ2:} Which design mechanisms contribute to decision-valid scientific execution?
\textbf{RQ3:} How consistently does HazardWeaver perform across language-model backbones and hazard-task settings?

\subsection{System Performance}
\label{sec:rq1}
We answer \textbf{RQ1} by comparing HazardWeaver with adapted agent systems and general control strategies. All methods use the same task inventory, Llama backbone, registered route catalog, and DCA evaluator. Table~\ref{tab:main-results} reports performance across hazard composition and route multiplicity.

\providecommand{\dcaci}[3]{%
  \ensuremath{#1^{\scriptscriptstyle +#2}_{\scriptscriptstyle -#3}}%
}

\begin{table}[t]
\centering
\caption{Comparison with agent baselines. Each cell reports DCA (\%) with the upper and lower offsets of its
Wilson 95\% confidence interval. Methods are ordered by overall DCA.}
\vspace{-0.3cm}
\label{tab:main-results}
{\setlength{\tabcolsep}{2.6pt}
\renewcommand{\arraystretch}{1.08}
\begin{tabular*}{\linewidth}{@{\extracolsep{\fill}}lccccc@{}}
\toprule
Method &
\shortstack{Single-hazard} &
\shortstack{Multi-hazard} &
\shortstack{One route} &
\shortstack{Multiple routes} &
\shortstack{Overall DCA} \\
\midrule
AIDE
& \dcaci{22.2}{9.7}{7.3}
& \dcaci{23.5}{13.3}{9.5}
& \dcaci{28.9}{14.5}{11.2}
& \dcaci{19.8}{9.1}{6.7}
& \dcaci{22.7}{7.6}{6.1} \\

AutoML-Agent
& \dcaci{26.7}{9.9}{8.1}
& \dcaci{27.4}{13.6}{10.3}
& \dcaci{24.4}{14.3}{10.2}
& \dcaci{28.1}{9.7}{8.0}
& \dcaci{26.9}{7.9}{6.6} \\

ReAct
& \dcaci{28.9}{10.1}{8.4}
& \dcaci{31.4}{13.6}{11.1}
& \dcaci{31.1}{14.6}{11.6}
& \dcaci{29.2}{9.7}{8.2}
& \dcaci{29.8}{8.0}{6.9} \\

Reflexion
& \dcaci{30.0}{10.1}{8.5}
& \dcaci{35.3}{13.7}{11.7}
& \dcaci{51.1}{13.9}{14.1}
& \dcaci{22.9}{9.4}{7.2}
& \dcaci{31.9}{8.1}{7.1} \\

MLE-STAR
& \dcaci{35.6}{10.3}{9.2}
& \dcaci{29.4}{13.6}{10.7}
& \dcaci{35.6}{14.6}{12.4}
& \dcaci{32.3}{9.9}{8.5}
& \dcaci{33.3}{8.2}{7.2} \\

DS-Agent
& \dcaci{33.3}{10.3}{8.9}
& \dcaci{33.3}{13.7}{11.3}
& \dcaci{40.0}{14.5}{13.0}
& \dcaci{30.2}{9.8}{8.3}
& \dcaci{33.3}{8.2}{7.2} \\

R\&D-Agent
& \dcaci{33.3}{10.3}{8.9}
& \dcaci{35.3}{13.7}{11.7}
& \dcaci{42.2}{14.5}{13.2}
& \dcaci{30.2}{9.8}{8.3}
& \dcaci{34.0}{8.2}{7.3} \\

Self-Consistency
& \dcaci{33.3}{10.3}{8.9}
& \dcaci{39.2}{13.7}{12.2}
& \dcaci{42.2}{14.5}{13.2}
& \dcaci{32.3}{9.9}{8.5}
& \dcaci{35.5}{8.1}{7.4} \\

DisasterBench-ToT
& \dcaci{35.6}{10.3}{9.2}
& \dcaci{35.3}{13.7}{11.7}
& \dcaci{42.2}{14.5}{13.2}
& \dcaci{32.3}{9.9}{8.5}
& \dcaci{35.5}{8.1}{7.4} \\

Plan-and-Solve
& \dcaci{41.1}{10.3}{9.6}
& \dcaci{41.2}{13.6}{12.4}
& \dcaci{57.8}{13.2}{14.5}
& \dcaci{33.3}{9.9}{8.6}
& \dcaci{41.1}{8.3}{7.7} \\

\midrule
\textbf{HazardWeaver}
& \dcaci{\textbf{91.1}}{4.3}{7.7}
& \dcaci{\textbf{86.3}}{6.9}{12.0}
& \dcaci{\textbf{80.0}}{9.1}{13.8}
& \dcaci{\textbf{93.8}}{3.3}{6.8}
& \dcaci{\textbf{89.4}}{4.1}{6.2} \\
\bottomrule
\vspace{-.9cm}
\end{tabular*}}
\end{table}

The comparison covers two complementary baseline families. AIDE, DS-Agent, R\&D-Agent, MLE-STAR, AutoML-Agent, and DisasterBench-ToT adapt representative agent-system designs to the benchmark interface. ReAct, Reflexion, Self-Consistency, and Plan-and-Solve evaluate alternative control strategies with the same backbone and scientific tools as HazardWeaver. HazardWeaver reaches 89.4\% overall DCA, compared with 41.1\% for the strongest baseline. The difference is most pronounced when several routes are available: Plan-and-Solve performs well with one route but declines sharply under multiple-route choice, while HazardWeaver rises to 93.8\%. Iterative refinement, retrieval, search, reflection, and explicit planning therefore provide limited benefit when the controller must identify which scientific workflow is valid. HazardWeaver remains strong across both hazard compositions and route regimes, indicating that its advantage comes from structured route eligibility rather than a particular task subset.

\subsection{Component and Mechanism Analysis}
\label{sec:rq2}

\textbf{RQ2} examines whether HazardWeaver benefits from explicitly representing scientific eligibility and maintaining it as execution proceeds. We test these two design choices separately. The first experiment removes the evidence-linked
conditions supplied by HKC or the typed compatibility constraints supplied by HCG. The second disables route updating on the subset of tasks for which eligibility can change during execution.

\begin{wrapfigure}[15]{r}{0.38\linewidth}
\vspace{-.4cm}
\centering
\includegraphics[width=\linewidth]{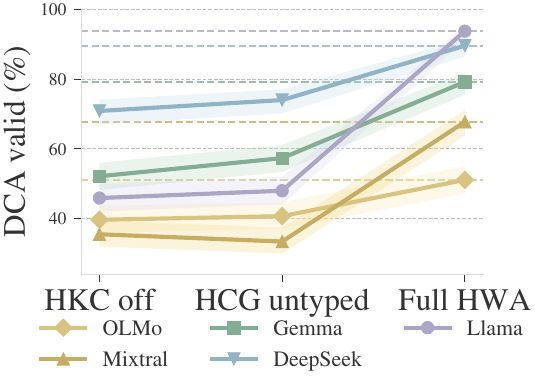}
\vspace{-.75cm}
\caption{Scientific-constraint ablation results. DCA (\%) on the 96 multiple-route instances. Shaded bands indicate Wilson 95\% interval.}
\label{fig:rq2-constraints}
\end{wrapfigure}
\textbf{Scientific constraints.}
Figure~\ref{fig:rq2-constraints} shows the same directional result across all five backbones: Full HazardWeaver outperforms both ablated variants. The effect is strongest for policies that can otherwise execute the selected workflow reliably. 
On Llama, full DCA reaches 93.8\%, while removing HKC conditions or HCG typing reduces it to 45.8\% and 47.9\%. DeepSeek exhibits the same pattern at a smaller scale. The weaker policies retain less performance overall, but neither ablation improves any backbone. These results distinguish scientific route selection from ordinary tool use. HKC removes routes whose assumptions do not match the task context, while HCG prevents compositions whose inputs and outputs are incompatible. Without either constraint, access to the correct scientific capabilities is insufficient for selecting a valid workflow.

\providecommand{\dcaci}[3]{%
  \ensuremath{#1^{\scriptscriptstyle +#2}_{\scriptscriptstyle -#3}}%
}
\begin{wraptable}[12]{r}{0.38\linewidth}
% \vspace{-.01cm}
\centering
\caption{Route update ablation. DCA (\%) on 40 applicable instances.}
\vspace{-0.4cm}
\label{tab:rq2-route-update}
\setlength{\tabcolsep}{2.2pt}
\renewcommand{\arraystretch}{1.12}
\begin{tabular*}{\linewidth}
{@{\extracolsep{\fill}}lll@{}}
\toprule
Backbone & Fixed & Updated \\
\midrule
Mixtral
& \dcaci{5.0}{11.5}{3.6}
& \textbf{\dcaci{15.0}{14.1}{7.9}} \\
OLMo
& \dcaci{17.5}{14.4}{8.7}
& \textbf{\dcaci{25.0}{15.2}{10.8}} \\
DeepSeek
& \dcaci{22.5}{15.0}{10.2}
& \textbf{\dcaci{35.0}{15.5}{12.9}} \\
Gemma
& \dcaci{20.0}{14.8}{9.5}
& \textbf{\dcaci{42.5}{15.3}{14.0}} \\
Llama
& \dcaci{15.0}{14.1}{7.9}
& \textbf{\dcaci{90.0}{6.0}{13.0}} \\
\midrule
Pooled
& {16.0}
& \textbf{{41.5}} \\
\bottomrule
\end{tabular*}
\vspace{-0.9em}
\end{wraptable}
\textbf{Route updating.} Table~\ref{tab:rq2-route-update} isolates the second design choice on the 40 benchmark instances where route eligibility may change during execution. Recomputing eligibility improves every backbone, raising pooled DCA from 16.0\% to 41.5\%. The gain is largest for Llama, whose stronger policy can exploit the revised route set once an earlier workflow becomes ineligible. The consistent direction across all five backbones shows that this benefit does not depend on one model family. A capable controller still requires an up-to-date decision space, since a fixed eligible set retains invalidated routes.
Overall, reliable scientific route decisions require evidence-linked applicability conditions, typed capability composition, and continued maintenance of the eligible route set.

\subsection{Generalization Across Backbones and Task Structures}
\label{sec:rq3}

\begin{table}[t]
\centering
\caption{Cross-backbone performance of Full HazardWeaver. DCA is the headline metric, and Macro DCA assigns equal weight to the 11 tracks. $E_q$ and $V_q$ report the corresponding marginal artifact- and trajectory-validity rates under the same sealed evaluator.}
\vspace{-0.3cm}
\label{tab:rq3-main}
\setlength{\tabcolsep}{5.0pt}
\renewcommand{\arraystretch}{1.10}
\begin{tabular*}{\linewidth}{@{\extracolsep{\fill}}lcccc@{}}
\toprule
Backbone
& DCA
& Macro DCA
& Artifact validity ($E_q$)
& Trajectory validity ($V_q$) \\
\midrule
OLMo
& 60.3 [52.0, 68.0]
& 61.9 [54.4, 71.2]
& 78.0 [70.5, 84.1]
& 91.5 [85.7, 95.1] \\
Mixtral
& 70.2 [62.2, 77.1]
& 70.8 [57.6, 83.9]
& 79.4 [72.0, 85.3]
& 92.2 [86.6, 95.6] \\
Gemma
& 79.4 [72.0, 85.3]
& 78.6 [65.2, 91.5]
& 86.5 [79.9, 91.2]
& 92.9 [87.4, 96.1] \\
DeepSeek
& 86.5 [79.9, 91.2]
& 86.6 [77.2, 94.3]
& 90.8 [84.9, 94.5]
& 92.2 [86.6, 95.6] \\
Llama
& \textbf{89.4} [83.2, 93.5]
& \textbf{89.5} [79.7, 97.0]
& \textbf{92.2} [86.6, 95.6]
& \textbf{95.0} [90.1, 97.6] \\ 
\bottomrule
\end{tabular*}
\vspace{-0.3cm}
\end{table}

\textbf{RQ3} examines whether HazardWeaver remains effective when its policy model changes and which task structures expose the remaining differences. We hold the benchmark, HKC conditions, HCG capabilities, execution interface, and evaluator fixed, and replace only the language-model backbone.

\textbf{Cross-backbone effectiveness.} Table~\ref{tab:rq3-main} shows that HazardWeaver remains effective across all five policy backbones, with overall DCA ranging from 60.3\% to 89.4\%. Llama and DeepSeek achieve the strongest results, while Gemma retains nearly 80\% DCA with a smaller model. Macro DCA closely follows the overall ordering, showing that the ranking is balanced by all tracks. The marginal validity rates sharpen this result: artifact validity broadly follows DCA, whereas trajectory validity remains consistently high across backbones. The remaining cross-backbone separation is therefore concentrated in producing the required scientific artifact and selecting the valid scientific route.

% \vspace{0.3cm}
\begin{figure}[t]
\centering
\begin{subfigure}[t]{0.37\linewidth}
    \centering
    \includegraphics[
        height=3.25cm,
        width=\linewidth,
        keepaspectratio
    ]{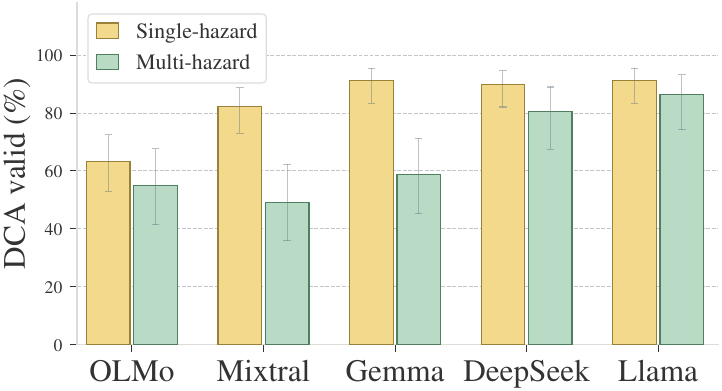}
    \caption{Hazard composition.}
    \label{fig:rq3-composition}
\end{subfigure}
\hfill
\begin{subfigure}[t]{0.37\linewidth}
    \centering
    \includegraphics[
        height=3.25cm,
        width=\linewidth,
        keepaspectratio
    ]{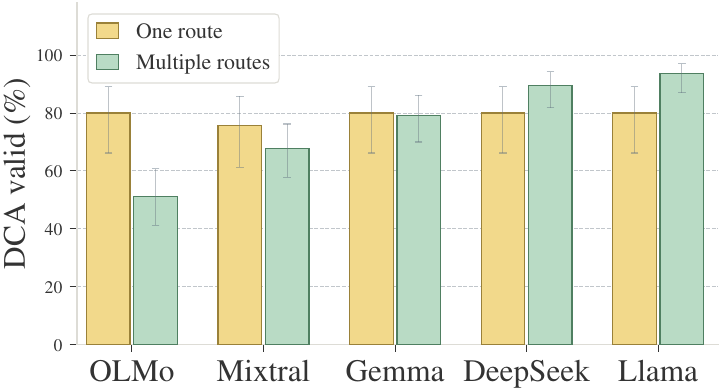}
    \caption{Route multiplicity.}
    \label{fig:rq3-routes}
\end{subfigure}
\hfill
\begin{subfigure}[t]{0.24\linewidth}
    \centering
    \includegraphics[
        height=3.25cm,
        width=\linewidth,
        keepaspectratio
    ]{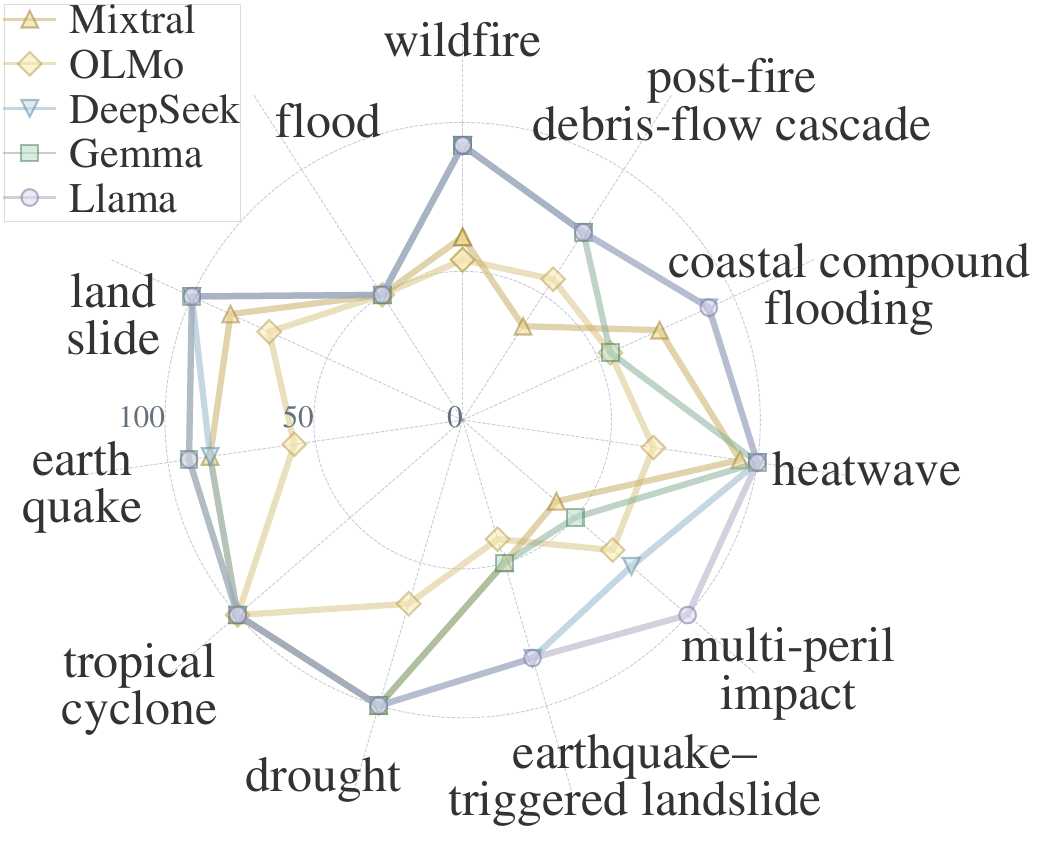}
    \caption{Track-level coverage.}
    \label{fig:rq3-tracks}
\end{subfigure}
\vspace{-0.3cm}
\caption{Cross-backbone behavior of Full HazardWeaver. Panels report DCA by (a) hazard composition, (b) route multiplicity, and (c) benchmark track.}
\label{fig:rq3-breakdown}
\vspace{-0.6cm}
\end{figure}

\textbf{Hazard composition.} Figure~\ref{fig:rq3-composition} shows that the strongest policies preserve most of their performance when tasks move from single-hazard analysis to multi-hazard composition. Llama and DeepSeek remain above 80\% on multi-hazard tasks, whereas the gap widens for Mixtral and Gemma. Scientific composition therefore provides a more discriminating test of the policy backbone than isolated hazard analysis, since the agent must carry intermediate state across stages and select a compatible downstream workflow while preserving scientific validity throughout the composed analysis.

\textbf{Route multiplicity.} Figure~\ref{fig:rq3-routes} reveals a sharper difference across route regimes. Performance on the one-route subset is tightly grouped across the five backbones, whereas the multiple-route subset produces substantially wider separation. Llama and DeepSeek achieve higher DCA on the multiple-route subset, while the weaker policies exploit the available alternatives less consistently. Route-rich tasks therefore provide the clearest distinction between policy backbones within the same HazardWeaver architecture. This descriptive pattern complements Section~\ref{sec:rq2}, where controlled ablations establish the contribution of the route-eligibility mechanisms themselves.

\textbf{Track-level coverage.} Figure~\ref{fig:rq3-tracks} confirms that the aggregate results reflect broad coverage rather than success on one hazard family. The strongest backbones form consistently broad profiles across single- and multi-hazard tracks. Flood remains the clearest shared bottleneck, while compound tracks such as multi-peril impact and earthquake-triggered landslide separate the policy backbones more strongly.
These differences identify where future improvements should focus: transferring scientific state across stages and choosing among competing composed workflows.

% Overall, RQ3 shows that HazardWeaver's route-selection architecture transfers across five materially different language-model backbones. Stronger policies extract greater benefit from the same scientific structure on multi-hazard and multiple-route tasks, while the shared framework preserves substantial decision-valid performance even with the weaker backbones.

\section{Related Work}
\label{sec:related}

\noindent\textbf{Scientific Route Grounding.} Hazard science provides specialized empirical, physical, statistical, and learned models for single- and multi-hazard analysis, such as fire-area simulators and inertial shallow-water flood solvers \citep{finney_farsite_1998,bates_simple_2010,xu2026nationwide}. These models encode scientific assumptions through equations, required observations, calibration regimes, and validity ranges, while scientific workflow research represents executable operations through interfaces, dependencies, and capability graphs \citep{ahn_can_2022,zhai_hyperagent_2026}. DisasterBench evaluates planning under typed tool interfaces \citep{chen_disasterbench_2026}, and DORA extends disaster agents to heterogeneous geospatial and learned tools with expert-authored tasks and replayable trajectories \citep{wang_can_2026}. These systems establish capable execution environments, but selecting among methods that target the same output also depends on variable definitions, units, spatial and temporal support, resolution, uncertainty, provenance, and operating scale \citep{baumann_analysisreadiness_2024,lee_machine_2025}. Multi-hazard analysis compounds this requirement, since the output of one hazard model becomes an input condition of the next \citep{kappes_challenges_2012}. Existing workflow representations capture executable dependencies well, while task-conditioned scientific applicability remains largely external to their route structures. \textsc{HazardWeaver} connects these layers by grounding alternative route families in evidence-linked conditions and composing them through typed executable capabilities.

\noindent\textbf{State-Dependent Route Decision.} Automated machine learning formulates estimator and pipeline selection as search over a fixed target and evaluation objective \citep{akiba_optuna_2019,wang_flaml_2021}. Recent agents extend this process to executable code, iterative experiment refinement, and data-science workflows \citep{shinn_reflexion_2023,trirat_automlagent_2025,nam_mlestar_2025,huang_mlagentbench_2024,cheng2026latentrouter}, while richer planning systems organize available tools before execution \citep{liu_llm+p_2023,kim_llm_2024,rana_sayplan_2023}. Hazard analysis additionally requires route eligibility to follow the current scientific state because new evidence, intermediate artifacts, and capability outcomes such as tool failures or missing inputs can change which workflows remain applicable. Related benchmarks evaluate planning, multi-step trajectories, and task outcomes \citep{valmeekam_planbench_2023,trivedi_appworld_2024,lu_toolsandbox_2025,pothuri_gisagentbench_2026,yao_$t$bench_2024,chan_mlebench_2024}, including compositional and changing environments \citep{wang_mcpbench_2025,yoa_static_2026,zeng_futurex_2025}, and recent abstention benchmarks study whether agents can recognize when to stop acting \citep{liu_agentabstain_2026}. Route-plural scientific analysis further requires relating an executed workflow to the state in which it was selected while accommodating multiple registered solutions, and judging abstention by whether any usable route remains reachable. \textsc{HazardWeaver} addresses this problem by recomputing eligible scientific routes after every executed step and evaluating the resulting artifact, the validity of the committed route, and the justification of any abstention.
\section{Conclusion}
\label{sec:conclusion}
We introduced state-dependent scientific route selection and \textsc{HazardWeaver}, which grounds applicability in literature evidence, composes models and tools through typed capability relations, and updates eligible workflows as analysis evolves. On 141 sealed instances spanning seven single-hazard domains and four multi-hazard classes, HazardWeaver achieves 89.4\% DCA versus 41.1\% for the strongest baseline, with the largest gains on multiple-route tasks. Ablations and evaluation across five backbones confirm the contributions and transferability of its route-selection architecture.

% We introduced state-dependent scientific route selection as a central decision problem for hazard-analysis agents and presented \textsc{HazardWeaver}, which grounds applicability in literature evidence, composes scientific models and tools through typed capability relations, and updates eligible workflows as the analysis state evolves. The Hazard Weaver Benchmark evaluates this process on 141 sealed instances across seven single-hazard domains and four multi-hazard interaction classes. HazardWeaver achieves 89.4\% Decision-Constrained Accuracy, compared with 41.1\% for the strongest representative agent baseline, with its largest advantage on tasks offering multiple scientific routes. Ablating evidence grounding, capability typing, or route updating consistently degrades performance, while evaluation across five backbones confirms that the architecture transfers beyond a single policy model. These results establish explicit, state-aware scientific route selection as a practical foundation for reliable model-backed hazard analysis.

\clearpage
% \subsection*{AI use statement}

% Generative AI tools were used to assist with language editing and improving the clarity and readability of the manuscript. All AI-assisted content was carefully reviewed and revised by the authors. The authors take full responsibility for the final content of this work.

% \section*{Ethics Statement}

% The authors have read and adhere to the ICLR Code of Ethics. This work involves no human subjects or personally identifiable information and uses only public scientific data. HazardWeaver is a research prototype and has not been validated for operational forecasting or early warning. Provenance records make its outputs auditable, and abstention withholds outputs when no valid route remains, reducing the risk of confident but invalid analyses. Operational use would require local calibration, expert review, and human oversight.

% \subsection*{Reproducibility statement}

% We provide detailed descriptions of the proposed method, experimental setup, datasets, evaluation protocols, and implementation details in the main paper and the appendix. Additional implementation details are available in the GitHub repository linked in the abstract.

\bibliography{references}

@article{cheng2026latentrouter,
  title   = {{LatentRouter}: Can We Choose the Right Multimodal Model Before Seeing Its Answer?},
  author  = {Cheng, Xueqi and Dong, Yushun},
  journal = {arXiv preprint arXiv:2605.11301},
  year    = {2026},
  doi     = {10.48550/arXiv.2605.11301},
  url     = {https://arxiv.org/abs/2605.11301}
}

@article{xu2026nationwide,
  title   = {A Nationwide Benchmark for Wildfire Initial Attack Failure Prediction with Public Environmental Data},
  author  = {Xu, Runyang and Cheng, Xueqi and Dong, Yushun},
  journal = {arXiv preprint arXiv:2606.15529},
  year    = {2026},
  doi     = {10.48550/arXiv.2606.15529},
  url     = {https://arxiv.org/abs/2606.15529}
}

@inproceedings{jaber2026autoclimds,
  title     = {{AutoClimDS}: Climate Data Science Agentic AI---A Knowledge Graph Is All You Need},
  author    = {Jaber, Ahmed and Zhu, Wangshu and Roy, Ayon and
               Jayavelu, Karthick and Downes, Justin and Mohamed, Sameer and
               Agonafir, Candace and Hawkins, Linnia and Zheng, Tian},
  booktitle = {2026 IEEE Conference on Artificial Intelligence (CAI)},
  pages     = {1057--1063},
  year      = {2026},
  doi       = {10.1109/CAI68641.2026.11536364},
  url       = {https://doi.org/10.1109/CAI68641.2026.11536364}
}

@misc{ahn_can_2022,
  title = {Do {{As I Can}}, {{Not As I Say}}: {{Grounding Language}} in {{Robotic Affordances}}},
  shorttitle = {Do {{As I Can}}, {{Not As I Say}}},
  author = {Ahn, Michael and Brohan, Anthony and Brown, Noah and Chebotar, Yevgen and Cortes, Omar and David, Byron and Finn, Chelsea and Fu, Chuyuan and Gopalakrishnan, Keerthana and Hausman, Karol and Herzog, Alex and Ho, Daniel and Hsu, Jasmine and Ibarz, Julian and Ichter, Brian and Irpan, Alex and Jang, Eric and Ruano, Rosario Jauregui and Jeffrey, Kyle and Jesmonth, Sally and Joshi, Nikhil J. and Julian, Ryan and Kalashnikov, Dmitry and Kuang, Yuheng and Lee, Kuang-Huei and Levine, Sergey and Lu, Yao and Luu, Linda and Parada, Carolina and Pastor, Peter and Quiambao, Jornell and Rao, Kanishka and Rettinghouse, Jarek and Reyes, Diego and Sermanet, Pierre and Sievers, Nicolas and Tan, Clayton and Toshev, Alexander and Vanhoucke, Vincent and Xia, Fei and Xiao, Ted and Xu, Peng and Xu, Sichun and Yan, Mengyuan and Zeng, Andy},
  year = 2022,
  month = aug,
  number = {arXiv:2204.01691},
  eprint = {2204.01691},
  primaryclass = {cs.RO},
  publisher = {arXiv},
  doi = {10.48550/arXiv.2204.01691},
  urldate = {2026-09-15},
  archiveprefix = {arXiv}
}

@inproceedings{akiba_optuna_2019,
  title = {Optuna: {{A Next-generation Hyperparameter Optimization Framework}}},
  shorttitle = {Optuna},
  booktitle = {Proceedings of the 25th {{ACM SIGKDD International Conference}} on {{Knowledge Discovery}} \& {{Data Mining}}},
  author = {Akiba, Takuya and Sano, Shotaro and Yanase, Toshihiko and Ohta, Takeru and Koyama, Masanori},
  year = 2019,
  month = jul,
  pages = {2623--2631},
  publisher = {ACM},
  address = {Anchorage AK USA},
  doi = {10.1145/3292500.3330701},
  urldate = {2026-08-21},
  isbn = {978-1-4503-6201-6},
  langid = {english}
}

@article{bates_simple_2010,
  title = {A Simple Inertial Formulation of the Shallow Water Equations for Efficient Two-Dimensional Flood Inundation Modelling},
  author = {Bates, Paul D. and Horritt, Matthew S. and Fewtrell, Timothy J.},
  year = 2010,
  month = jun,
  journal = {Journal of Hydrology},
  volume = {387},
  number = {1-2},
  pages = {33--45},
  issn = {00221694},
  doi = {10.1016/j.jhydrol.2010.03.027},
  urldate = {2026-08-21},
  copyright = {https://www.elsevier.com/tdm/userlicense/1.0/},
  langid = {english}
}

@article{baumann_analysisreadiness_2024,
  title = {On the Analysis-Readiness of Spatio-Temporal {{Earth}} Data and Suggestions for Its Enhancement},
  author = {Baumann, Peter},
  year = 2024,
  month = may,
  journal = {Environmental Modelling \& Software},
  volume = {176},
  pages = {106017},
  issn = {13648152},
  doi = {10.1016/j.envsoft.2024.106017},
  urldate = {2026-08-21},
  langid = {english}
}

@article{boiko_autonomous_2023a,
  title = {Autonomous Chemical Research with Large Language Models},
  author = {Boiko, Daniil A. and MacKnight, Robert and Kline, Ben and Gomes, Gabe},
  year = 2023,
  month = dec,
  journal = {Nature},
  volume = {624},
  number = {7992},
  pages = {570--578},
  publisher = {Nature Publishing Group},
  issn = {1476-4687},
  doi = {10.1038/s41586-023-06792-0},
  urldate = {2026-09-15},
  copyright = {2023 The Author(s)},
  langid = {english}
}

@misc{chen_disasterbench_2026,
  title = {{{DisasterBench}}: {{Benchmarking LLM Planning}} under {{Typed Tool Interface Constraints}}},
  shorttitle = {{{DisasterBench}}},
  author = {Chen, Zhitong and Yin, Kai and Zhang, Weifeng and Wang, Zhiyuan and Dong, Xiangjue and Liu, Chengkai and Liu, Zhewei and Xiao, Yiming and Mostafavi, Ali and Caverlee, James},
  year = 2026,
  month = may,
  number = {arXiv:2605.27957},
  eprint = {2605.27957},
  primaryclass = {cs.CL},
  publisher = {arXiv},
  doi = {10.48550/arXiv.2605.27957},
  urldate = {2026-09-15},
  archiveprefix = {arXiv}
}

@article{deepseek_deepseekv41flash_2026,
  title = {{{DeepSeek-V4}}.1-{{Flash}}: {{Pushing}} the {{Limits}} of {{KV Cache Compression}}},
  shorttitle = {{{DeepSeek-V4}}.1-{{Flash}}},
  author = {{Deepseek}},
  year = 2026,
  month = sep,
  urldate = {2026-09-15},
  langid = {english}
}

@techreport{finney_farsite_1998,
  title = {{{FARSITE}}: {{Fire Area Simulator-model}} Development and Evaluation},
  shorttitle = {{{FARSITE}}},
  author = {Finney, Mark A.},
  year = 1998,
  number = {RMRS-RP-4},
  pages = {RMRS-RP-4},
  address = {Ft. Collins, CO},
  institution = {U.S. Department of Agriculture, Forest Service, Rocky Mountain Research Station},
  doi = {10.2737/RMRS-RP-4},
  urldate = {2026-08-21},
  langid = {english}
}

@article{gil_wings_2011,
  title = {Wings: {{Intelligent Workflow-Based Design}} of {{Computational Experiments}}},
  shorttitle = {Wings},
  author = {Gil, Yolanda and Ratnakar, Varun and Kim, Jihie and {Gonzalez-Calero}, Pedro and Groth, Paul and Moody, Joshua and Deelman, Ewa},
  year = 2011,
  month = jan,
  journal = {IEEE Intelligent Systems},
  volume = {26},
  number = {1},
  pages = {62--72},
  issn = {1941-1294},
  doi = {10.1109/MIS.2010.9},
  urldate = {2026-09-15}
}

@article{gill_reviewing_2014a,
  title = {Reviewing and Visualizing the Interactions of Natural Hazards},
  author = {Gill, Joel C. and Malamud, Bruce D.},
  year = 2014,
  journal = {Reviews of Geophysics},
  volume = {52},
  number = {4},
  pages = {680--722},
  issn = {1944-9208},
  doi = {10.1002/2013RG000445},
  urldate = {2026-09-15},
  copyright = {\copyright 2014. The Authors.},
  langid = {english}
}

@article{gorr_improved_2026,
  title = {An Improved Empirical Model for Predicting Postfire Debris-Flow Volume in the Western {{United States}}},
  author = {Gorr, Alexander N. and Rengers, Francis K. and Barnhart, Katherine R. and Thomas, Matthew A. and Kean, Jason W.},
  year = 2026,
  month = may,
  journal = {Natural Hazards and Earth System Sciences},
  volume = {26},
  number = {5},
  pages = {2111--2132},
  publisher = {Copernicus GmbH},
  issn = {1561-8633},
  doi = {10.5194/nhess-26-2111-2026},
  urldate = {2026-09-15},
  langid = {english}
}

@misc{grattafiori_llama_2024,
  title = {The {{Llama}} 3 {{Herd}} of {{Models}}},
  author = {Grattafiori, Aaron and Dubey, Abhimanyu and others},
  year = 2024,
  month = nov,
  number = {arXiv:2407.21783},
  eprint = {2407.21783},
  primaryclass = {cs.AI},
  publisher = {arXiv},
  doi = {10.48550/arXiv.2407.21783},
  urldate = {2026-09-15},
  archiveprefix = {arXiv}
}

@misc{guo_dsagent_2024,
  title = {{{DS-Agent}}: {{Automated Data Science}} by {{Empowering Large Language Models}} with {{Case-Based Reasoning}}},
  shorttitle = {{{DS-Agent}}},
  author = {Guo, Siyuan and Deng, Cheng and Wen, Ying and Chen, Hechang and Chang, Yi and Wang, Jun},
  year = 2024,
  month = may,
  number = {arXiv:2402.17453},
  eprint = {2402.17453},
  primaryclass = {cs.LG},
  publisher = {arXiv},
  doi = {10.48550/arXiv.2402.17453},
  urldate = {2026-09-15},
  archiveprefix = {arXiv}
}

@article{huot_next_2022,
  title = {Next {{Day Wildfire Spread}}: {{A Machine Learning Data Set}} to {{Predict Wildfire Spreading}} from {{Remote-Sensing Data}}},
  shorttitle = {Next {{Day Wildfire Spread}}},
  author = {Huot, Fantine and Hu, R. Lily and Goyal, Nita and Sankar, Tharun and Ihme, Matthias and Chen, Yi-Fan},
  year = 2022,
  journal = {IEEE Transactions on Geoscience and Remote Sensing},
  volume = {60},
  eprint = {2112.02447},
  primaryclass = {cs.CV},
  pages = {1--13},
  issn = {0196-2892, 1558-0644},
  doi = {10.1109/TGRS.2022.3192974},
  urldate = {2026-06-22},
  archiveprefix = {arXiv}
}

@inproceedings{jain_scirex_2020,
  title = {{{SciREX}}: {{A Challenge Dataset}} for {{Document-Level Information Extraction}}},
  shorttitle = {{{SciREX}}},
  booktitle = {Proceedings of the 58th {{Annual Meeting}} of the {{Association}} for {{Computational Linguistics}}},
  author = {Jain, Sarthak and {van Zuylen}, Madeleine and Hajishirzi, Hannaneh and Beltagy, Iz},
  editor = {Jurafsky, Dan and Chai, Joyce and Schluter, Natalie and Tetreault, Joel},
  year = 2020,
  month = jul,
  pages = {7506--7516},
  publisher = {Association for Computational Linguistics},
  address = {Online},
  doi = {10.18653/v1/2020.acl-main.670},
  urldate = {2026-09-15}
}

@misc{jakubik_foundation_2023,
  title = {Foundation {{Models}} for {{Generalist Geospatial Artificial Intelligence}}},
  author = {Jakubik, Johannes and Roy, Sujit and Phillips, C. E. and Fraccaro, Paolo and Godwin, Denys and Zadrozny, Bianca and Szwarcman, Daniela and Gomes, Carlos and Nyirjesy, Gabby and Edwards, Blair and Kimura, Daiki and Simumba, Naomi and Chu, Linsong and Mukkavilli, S. Karthik and Lambhate, Devyani and Das, Kamal and Bangalore, Ranjini and Oliveira, Dario and Muszynski, Michal and Ankur, Kumar and Ramasubramanian, Muthukumaran and Gurung, Iksha and Khallaghi, Sam and Hanxi and Li and Cecil, Michael and Ahmadi, Maryam and Kordi, Fatemeh and Alemohammad, Hamed and Maskey, Manil and Ganti, Raghu and Weldemariam, Kommy and Ramachandran, Rahul},
  year = 2023,
  month = nov,
  number = {arXiv:2310.18660},
  eprint = {2310.18660},
  primaryclass = {cs.CV},
  publisher = {arXiv},
  doi = {10.48550/arXiv.2310.18660},
  urldate = {2026-09-15},
  archiveprefix = {arXiv}
}

@misc{jiang_aide_2025,
  title = {{{AIDE}}: {{AI-Driven Exploration}} in the {{Space}} of {{Code}}},
  shorttitle = {{{AIDE}}},
  author = {Jiang, Zhengyao and Schmidt, Dominik and Srikanth, Dhruv and Xu, Dixing and Kaplan, Ian and Jacenko, Deniss and Wu, Yuxiang},
  year = 2025,
  publisher = {arXiv},
  doi = {10.48550/ARXIV.2502.13138},
  urldate = {2026-08-21},
  copyright = {Creative Commons Attribution 4.0 International}
}

@misc{jiang_mixtral_2024,
  title = {Mixtral of {{Experts}}},
  author = {Jiang, Albert Q. and Sablayrolles, Alexandre and Roux, Antoine and Mensch, Arthur and Savary, Blanche and Bamford, Chris and Chaplot, Devendra Singh and de las Casas, Diego and Hanna, Emma Bou and Bressand, Florian and Lengyel, Gianna and Bour, Guillaume and Lample, Guillaume and Lavaud, L{\'e}lio Renard and Saulnier, Lucile and Lachaux, Marie-Anne and Stock, Pierre and Subramanian, Sandeep and Yang, Sophia and Antoniak, Szymon and Scao, Teven Le and Gervet, Th{\'e}ophile and Lavril, Thibaut and Wang, Thomas and Lacroix, Timoth{\'e}e and Sayed, William El},
  year = 2024,
  month = jan,
  number = {arXiv:2401.04088},
  eprint = {2401.04088},
  primaryclass = {cs.LG},
  publisher = {arXiv},
  doi = {10.48550/arXiv.2401.04088},
  urldate = {2026-09-15},
  archiveprefix = {arXiv}
}

@article{kappes_challenges_2012,
  title = {Challenges of Analyzing Multi-Hazard Risk: A Review},
  shorttitle = {Challenges of Analyzing Multi-Hazard Risk},
  author = {Kappes, Melanie S. and Keiler, Margreth and Von Elverfeldt, Kirsten and Glade, Thomas},
  year = 2012,
  month = nov,
  journal = {Natural Hazards},
  volume = {64},
  number = {2},
  pages = {1925--1958},
  issn = {0921-030X, 1573-0840},
  doi = {10.1007/s11069-012-0294-2},
  urldate = {2026-08-21},
  copyright = {http://www.springer.com/tdm},
  langid = {english}
}

@article{karpatne_theoryguided_2017,
  title = {Theory-Guided {{Data Science}}: {{A New Paradigm}} for {{Scientific Discovery}} from {{Data}}},
  shorttitle = {Theory-Guided {{Data Science}}},
  author = {Karpatne, Anuj and Atluri, Gowtham and Faghmous, James and Steinbach, Michael and Banerjee, Arindam and Ganguly, Auroop and Shekhar, Shashi and Samatova, Nagiza and Kumar, Vipin},
  year = 2017,
  month = oct,
  journal = {IEEE Transactions on Knowledge and Data Engineering},
  volume = {29},
  number = {10},
  eprint = {1612.08544},
  primaryclass = {cs.LG},
  pages = {2318--2331},
  issn = {1041-4347},
  doi = {10.1109/TKDE.2017.2720168},
  urldate = {2026-09-15},
  archiveprefix = {arXiv}
}

@misc{kim_llm_2024,
  title = {An {{LLM Compiler}} for {{Parallel Function Calling}}},
  author = {Kim, Sehoon and Moon, Suhong and Tabrizi, Ryan and Lee, Nicholas and Mahoney, Michael W. and Keutzer, Kurt and Gholami, Amir},
  year = 2024,
  month = jun,
  number = {arXiv:2312.04511},
  eprint = {2312.04511},
  primaryclass = {cs.CL},
  publisher = {arXiv},
  doi = {10.48550/arXiv.2312.04511},
  urldate = {2026-09-15},
  archiveprefix = {arXiv}
}

@article{lee_machine_2025,
  title = {Machine Learning-Based Rainfall-Induced Landslide Susceptibility Model and Short-Term Early Warning Assessment in {{South Korea}}},
  author = {Lee, Sujong and Roh, Minwoo and Jo, Hyun-Woo and Joon, Kim and Lee, Woo-Kyun},
  year = 2025,
  month = aug,
  journal = {Landslides},
  volume = {22},
  number = {8},
  pages = {2809--2827},
  issn = {1612-510X, 1612-5118},
  doi = {10.1007/s10346-025-02513-y},
  urldate = {2026-08-21},
  langid = {english}
}

@misc{liu_agentabstain_2026,
  title = {{{AgentAbstain}}: {{Do LLM Agents Know When Not}} to {{Act}}?},
  shorttitle = {{{AgentAbstain}}},
  author = {Liu, Xun and Zhang, Yi Evie and Kasprova, Vira and Rabbani, Parisa and Zahraei, Pardis Sadat and Zhang, Tianyu and {Ebrahimpour-Boroojeny}, Ali and Chandrasekaran, Varun},
  year = 2026,
  month = jul,
  number = {arXiv:2607.10059},
  eprint = {2607.10059},
  primaryclass = {cs.AI},
  publisher = {arXiv},
  doi = {10.48550/arXiv.2607.10059},
  urldate = {2026-09-15},
  archiveprefix = {arXiv}
}

@misc{liu_llm+p_2023,
  title = {{{LLM}}+{{P}}: {{Empowering Large Language Models}} with {{Optimal Planning Proficiency}}},
  shorttitle = {{{LLM}}+{{P}}},
  author = {Liu, Bo and Jiang, Yuqian and Zhang, Xiaohan and Liu, Qiang and Zhang, Shiqi and Biswas, Joydeep and Stone, Peter},
  year = 2023,
  month = sep,
  number = {arXiv:2304.11477},
  eprint = {2304.11477},
  primaryclass = {cs.AI},
  publisher = {arXiv},
  doi = {10.48550/arXiv.2304.11477},
  urldate = {2026-09-15},
  archiveprefix = {arXiv}
}

@inproceedings{lu_toolsandbox_2025,
  title = {{{ToolSandbox}}: {{A Stateful}}, {{Conversational}}, {{Interactive Evaluation Benchmark}} for {{LLM Tool Use Capabilities}}},
  shorttitle = {{{ToolSandbox}}},
  booktitle = {Findings of the {{Association}} for {{Computational Linguistics}}: {{NAACL}} 2025},
  author = {Lu, Jiarui and Holleis, Thomas and Zhang, Yizhe and Aumayer, Bernhard and Nan, Feng and Bai, Haoping and Ma, Shuang and Ma, Shen and Li, Mengyu and Yin, Guoli and Wang, Zirui and Pang, Ruoming},
  editor = {Chiruzzo, Luis and Ritter, Alan and Wang, Lu},
  year = 2025,
  month = apr,
  pages = {1160--1183},
  publisher = {Association for Computational Linguistics},
  address = {Albuquerque, New Mexico},
  doi = {10.18653/v1/2025.findings-naacl.65},
  urldate = {2026-05-26},
  isbn = {979-8-89176-195-7}
}

@inproceedings{luan_multitask_2018,
  title = {Multi-{{Task Identification}} of {{Entities}}, {{Relations}}, and {{Coreference}} for {{Scientific Knowledge Graph Construction}}},
  booktitle = {Proceedings of the 2018 {{Conference}} on {{Empirical Methods}} in {{Natural Language Processing}}},
  author = {Luan, Yi and He, Luheng and Ostendorf, Mari and Hajishirzi, Hannaneh},
  editor = {Riloff, Ellen and Chiang, David and Hockenmaier, Julia and Tsujii, Jun'ichi},
  year = 2018,
  month = oct,
  pages = {3219--3232},
  publisher = {Association for Computational Linguistics},
  address = {Brussels, Belgium},
  doi = {10.18653/v1/D18-1360},
  urldate = {2026-09-15}
}

@article{m.bran_augmenting_2024a,
  title = {Augmenting Large Language Models with Chemistry Tools},
  author = {Bran, Andres M. and Cox, Sam and Schilter, Oliver and Baldassari, Carlo and White, Andrew D. and Schwaller, Philippe},
  year = 2024,
  month = may,
  journal = {Nature Machine Intelligence},
  volume = {6},
  number = {5},
  pages = {525--535},
  publisher = {Nature Publishing Group},
  issn = {2522-5839},
  doi = {10.1038/s42256-024-00832-8},
  urldate = {2026-09-15},
  copyright = {2024 The Author(s)},
  langid = {english}
}

@inproceedings{nam_mlestar_2025,
  title = {{{MLE-STAR}}: {{Machine Learning Engineering Agent}} via {{Search}} and {{Targeted Refinement}}},
  shorttitle = {{{MLE-STAR}}},
  booktitle = {Advances in {{Neural Information Processing Systems}} 38},
  author = {Nam, Jaehyun and Yoon, Jinsung and Chen, Jiefeng and Shin, Jinwoo and Arik, Sercan and Pfister, Tomas},
  year = 2025,
  pages = {129290--129310},
  publisher = {Neural Information Processing Systems Foundation, Inc. (NeurIPS)},
  address = {San Diego, California, USA and Mexico City, Mexico},
  doi = {10.52202/085713-3889},
  urldate = {2026-08-21},
  isbn = {979-8-3313-3827-5}
}

@article{nearing_global_2024,
  title = {Global Prediction of Extreme Floods in Ungauged Watersheds},
  author = {Nearing, Grey and Cohen, Deborah and Dube, Vusumuzi and Gauch, Martin and Gilon, Oren and Harrigan, Shaun and Hassidim, Avinatan and Klotz, Daniel and Kratzert, Frederik and Metzger, Asher and Nevo, Sella and Pappenberger, Florian and Prudhomme, Christel and Shalev, Guy and Shenzis, Shlomo and Tekalign, Tadele Yednkachw and Weitzner, Dana and Matias, Yossi},
  year = 2024,
  month = mar,
  journal = {Nature},
  volume = {627},
  number = {8004},
  pages = {559--563},
  issn = {0028-0836, 1476-4687},
  doi = {10.1038/s41586-024-07145-1},
  urldate = {2026-08-21},
  langid = {english}
}

@misc{olmo_olmo_2026,
  title = {Olmo 3},
  author = {Olmo, Team and Ettinger, Allyson and Bertsch, Amanda and Kuehl, Bailey and Graham, David and Heineman, David and Groeneveld, Dirk and Brahman, Faeze and Timbers, Finbarr and others},
  year = 2026,
  month = apr,
  number = {arXiv:2512.13961},
  eprint = {2512.13961},
  primaryclass = {cs.CL},
  publisher = {arXiv},
  doi = {10.48550/arXiv.2512.13961},
  urldate = {2026-09-15},
  archiveprefix = {arXiv}
}

@article{perry_current_1998,
  title = {Current Approaches to Modelling the Spread of Wildland Fire: A Review},
  shorttitle = {Current Approaches to Modelling the Spread of Wildland Fire},
  author = {Perry, G. L.W.},
  year = 1998,
  month = jun,
  journal = {Progress in Physical Geography: Earth and Environment},
  volume = {22},
  number = {2},
  pages = {222--245},
  publisher = {SAGE Publications Ltd},
  issn = {0309-1333},
  doi = {10.1177/030913339802200204},
  urldate = {2026-08-21},
  langid = {english}
}

@misc{pothuri_gisagentbench_2026,
  title = {{{GISAgentBench}}: {{A Practitioner-Sourced Benchmark}} for {{Evaluating LLM Agents}} on {{GIS Tasks}}},
  shorttitle = {{{GISAgentBench}}},
  author = {Pothuri, Abhinav and Jiang, Zhe and Xu, Zelin and Yang, Di},
  year = 2026,
  month = aug,
  number = {arXiv:2608.01645},
  eprint = {2608.01645},
  primaryclass = {cs.AI},
  publisher = {arXiv},
  doi = {10.48550/arXiv.2608.01645},
  urldate = {2026-08-21},
  archiveprefix = {arXiv}
}

@article{rana_sayplan_2023,
  title = {{{SayPlan}}: {{Grounding Large Language Models}} Using {{3D Scene Graphs}} for {{Scalable Robot Task Planning}}},
  author = {Rana, Krishan and Haviland, Jesse and Garg, Sourav and {Abou-Chakra}, Jad and Reid, Ian},
  year={2023},
  eprint={2307.06135},
  archivePrefix={arXiv},
  primaryClass={cs.RO},
  url={https://arxiv.org/abs/2307.06135}, 
  langid = {english}
}

@article{reichstein_deep_2019,
  title = {Deep Learning and Process Understanding for Data-Driven {{Earth}} System Science},
  author = {Reichstein, Markus and {Camps-Valls}, Gustau and Stevens, Bjorn and Jung, Martin and Denzler, Joachim and Carvalhais, Nuno and Prabhat},
  year = 2019,
  month = feb,
  journal = {Nature},
  volume = {566},
  number = {7743},
  pages = {195--204},
  publisher = {Nature Publishing Group},
  issn = {1476-4687},
  doi = {10.1038/s41586-019-0912-1},
  urldate = {2026-09-15},
  copyright = {2019 Springer Nature Limited},
  langid = {english}
}

@article{stewart_selection_2015,
  title = {Selection of {{Ground Motion Prediction Equations}} for the {{Global Earthquake Model}}},
  author = {Stewart, Jonathan P. and Douglas, John and Javanbarg, Mohammad and Bozorgnia, Yousef and Abrahamson, Norman A. and Boore, David M. and Campbell, Kenneth W. and Delavaud, Elise and Erdik, Mustafa and Stafford, Peter J.},
  year = 2015,
  month = feb,
  journal = {Earthquake Spectra},
  volume = {31},
  number = {1},
  pages = {19--45},
  issn = {8755-2930, 1944-8201},
  doi = {10.1193/013013EQS017M},
  urldate = {2026-08-21},
  copyright = {http://onlinelibrary.wiley.com/termsAndConditions\#vor},
  langid = {english}
}

@misc{wang_can_2026,
  title = {Can {{LLM Agents Respond}} to {{Disasters}}? {{Benchmarking Heterogeneous Geospatial Reasoning}} in {{Emergency Operations}}},
  shorttitle = {Can {{LLM Agents Respond}} to {{Disasters}}?},
  author = {Wang, Junjue and Xuan, Weihao and Qi, Heli and Dai, Pengyu and Liu, Kunyi and Chen, Hongruixuan and Zheng, Zhuo and Xia, Junshi and Ermon, Stefano and Yokoya, Naoto},
  year = 2026,
  publisher = {arXiv},
  doi = {10.48550/ARXIV.2605.11633},
  urldate = {2026-08-21},
  copyright = {Creative Commons Attribution Non Commercial Share Alike 4.0 International}
}

@inproceedings{wang_flaml_2021,
  title = {{{FLAML}}: {{A Fast}} and {{Lightweight AutoML Library}}},
  booktitle = {Proceedings of the {{Fourth Conference}} on {{Machine Learning}} and {{Systems}} ({{MLSys}} 2021)},
  author = {Wang, Chi and Wu, Qingyun and Weimer, Markus and Zhu, Erkang},
  year = 2021,
  volume = {3},
  pages = {434--447}
}

@misc{wang_mcpbench_2025,
  title = {{{MCP-Bench}}: {{Benchmarking Tool-Using LLM Agents}} with {{Complex Real-World Tasks}} via {{MCP Servers}}},
  shorttitle = {{{MCP-Bench}}},
  author = {Wang, Zhenting and Chang, Qi and Patel, Hemani and Biju, Shashank and Wu, Cheng-En and Liu, Quan and Ding, Aolin and Rezazadeh, Alireza and Shah, Ankit and Bao, Yujia and Siow, Eugene},
  year = 2025,
  month = aug,
  number = {arXiv:2508.20453},
  eprint = {2508.20453},
  primaryclass = {cs.CL},
  publisher = {arXiv},
  doi = {10.48550/arXiv.2508.20453},
  urldate = {2026-09-15},
  archiveprefix = {arXiv}
}

@article{wilson_probable_1927,
  title = {Probable {{Inference}}, the {{Law}} of {{Succession}}, and {{Statistical Inference}}},
  author = {Wilson, Edwin B.},
  year = 1927,
  month = jun,
  journal = {Journal of the American Statistical Association},
  volume = {22},
  number = {158},
  pages = {209--212},
  publisher = {Taylor \& Francis},
  issn = {0162-1459},
  doi = {10.1080/01621459.1927.10502953},
  urldate = {2026-09-15}
}

@misc{yang_rdagent_2025,
  title = {R\&{{D-Agent}}: {{An LLM-Agent Framework Towards Autonomous Data Science}}},
  shorttitle = {R\&{{D-Agent}}},
  author = {Yang, Xu and Yang, Xiao and Fang, Shikai and Zhang, Yifei and Wang, Jian and Xian, Bowen and Li, Qizheng and Li, Jingyuan and Xu, Minrui and Li, Yuante and Pan, Haoran and Zhang, Yuge and Liu, Weiqing and Shen, Yelong and Chen, Weizhu and Bian, Jiang},
  year = 2025,
  month = oct,
  number = {arXiv:2505.14738},
  eprint = {2505.14738},
  primaryclass = {cs.AI},
  publisher = {arXiv},
  doi = {10.48550/arXiv.2505.14738},
  urldate = {2026-09-15},
  archiveprefix = {arXiv}
}

@misc{yao_$t$bench_2024,
  title = {\${$\tau\$$}-Bench: {{A Benchmark}} for {{Tool-Agent-User Interaction}} in {{Real-World Domains}}},
  shorttitle = {\${$\tau\$$}-Bench},
  author = {Yao, Shunyu and Shinn, Noah and Razavi, Pedram and Narasimhan, Karthik},
  year = 2024,
  publisher = {arXiv},
  doi = {10.48550/ARXIV.2406.12045},
  urldate = {2026-08-21},
  copyright = {Creative Commons Attribution 4.0 International}
}

@misc{yao_react_2022,
  title = {{{ReAct}}: {{Synergizing Reasoning}} and {{Acting}} in {{Language Models}}},
  shorttitle = {{{ReAct}}},
  author = {Yao, Shunyu and Zhao, Jeffrey and Yu, Dian and Du, Nan and Shafran, Izhak and Narasimhan, Karthik and Cao, Yuan},
  year = 2022,
  publisher = {arXiv},
  doi = {10.48550/ARXIV.2210.03629},
  urldate = {2026-08-21},
  copyright = {Creative Commons Attribution 4.0 International}
}

@misc{yoa_static_2026,
  title = {From {{Static Benchmarks}} to {{Dynamic Protocol}}: {{Agent-Centric Text Anomaly Detection}} for {{Evaluating LLM Reasoning}}},
  shorttitle = {From {{Static Benchmarks}} to {{Dynamic Protocol}}},
  author = {Yoa, Seungdong and Yoon, Sanghyu and Yoon, Suhee and Kim, Dongmin and Sim, Ye Seul and Lee, Junhyun and Lim, Woohyung},
  year = 2026,
  month = feb,
  number = {arXiv:2602.23729},
  eprint = {2602.23729},
  primaryclass = {cs.CL},
  publisher = {arXiv},
  doi = {10.48550/arXiv.2602.23729},
  urldate = {2026-09-15},
  archiveprefix = {arXiv}
}

@misc{zeng_futurex_2025,
  title = {{{FutureX}}: {{An Advanced Live Benchmark}} for {{LLM Agents}} in {{Future Prediction}}},
  shorttitle = {{{FutureX}}},
  author = {Zeng, Zhiyuan and Liu, Jiashuo and Chen, Siyuan and He, Tianci and Liao, Yali and Tian, Yixiao and Wang, Jinpeng and Wang, Zaiyuan and Yang, Yang and Yin, Lingyue and Yin, Mingren and Zhu, Zhenwei and Cai, Tianle and Chen, Zehui and Chen, Jiecao and Du, Yantao and Gao, Xiang and Guo, Jiacheng and Hu, Liang and Jiao, Jianpeng and Li, Xiangsheng and Liu, Jingkai and Ni, Shuang and Wen, Zhoufutu and Zhang, Ge and Zhang, Kaiyuan and Zhou, Xin and Blanchet, Jose and Qiu, Xipeng and Wang, Mengdi and Huang, Wenhao},
  year = 2025,
  month = sep,
  number = {arXiv:2508.11987},
  eprint = {2508.11987},
  primaryclass = {cs.AI},
  publisher = {arXiv},
  doi = {10.48550/arXiv.2508.11987},
  urldate = {2026-09-15},
  archiveprefix = {arXiv}
}

@misc{zhai_hyperagent_2026,
  title = {{{HyperAgent}}: {{Planning}} and {{Acting}} over {{Tool-Schema Hypergraphs}} for {{Tool-Use LLM Agents}}},
  shorttitle = {{{HyperAgent}}},
  author = {Zhai, Zian and Tan, Xingyu and Zou, Gaowang and Wang, Xiaoyang and Zhang, Wenjie},
  year = 2026,
  month = jul,
  number = {arXiv:2608.02650},
  eprint = {2608.02650},
  primaryclass = {cs.AI},
  publisher = {arXiv},
  doi = {10.48550/arXiv.2608.02650},
  urldate = {2026-09-15},
  archiveprefix = {arXiv}
}

@article{zhu_phasenet_2018,
  title = {{{PhaseNet}}: {{A Deep-Neural-Network-Based Seismic Arrival Time Picking Method}}},
  shorttitle = {{{PhaseNet}}},
  author = {Zhu, Weiqiang and Beroza, Gregory C},
  year = 2018,
  month = oct,
  journal = {Geophysical Journal International},
  issn = {0956-540X, 1365-246X},
  doi = {10.1093/gji/ggy423},
  urldate = {2026-08-21},
  copyright = {https://academic.oup.com/journals/pages/open\_access/funder\_policies/chorus/standard\_publication\_model},
  langid = {english}
}

@article{zscheischler_future_2018,
  title = {Future Climate Risk from Compound Events},
  author = {Zscheischler, Jakob and Westra, Seth and {van den Hurk}, Bart J. J. M. and Seneviratne, Sonia I. and Ward, Philip J. and Pitman, Andy and AghaKouchak, Amir and Bresch, David N. and Leonard, Michael and Wahl, Thomas and Zhang, Xuebin},
  year = 2018,
  month = jun,
  journal = {Nature Climate Change},
  volume = {8},
  number = {6},
  pages = {469--477},
  publisher = {Nature Publishing Group},
  issn = {1758-6798},
  doi = {10.1038/s41558-018-0156-3},
  urldate = {2026-09-15},
  copyright = {2018 Springer Nature Limited},
  langid = {english}
}

@article{zscheischler_typology_2020,
  title = {A Typology of Compound Weather and Climate Events},
  author = {Zscheischler, Jakob and Martius, Olivia and Westra, Seth and Bevacqua, Emanuele and Raymond, Colin and Horton, Radley M. and {van den Hurk}, Bart and AghaKouchak, Amir and J{\'e}z{\'e}quel, Agla{\'e} and Mahecha, Miguel D. and Maraun, Douglas and Ramos, Alexandre M. and Ridder, Nina N. and Thiery, Wim and Vignotto, Edoardo},
  year = 2020,
  month = jul,
  journal = {Nature Reviews Earth \& Environment},
  volume = {1},
  number = {7},
  pages = {333--347},
  publisher = {Nature Publishing Group},
  issn = {2662-138X},
  doi = {10.1038/s43017-020-0060-z},
  urldate = {2026-09-15},
  copyright = {2020 Springer Nature Limited},
  langid = {english}
}

@article{allstadt_us_2022,
  title = {The {{US Geological Survey}} Ground Failure Product: {{Near-real-time}} Estimates of Earthquake-Triggered Landslides and Liquefaction},
  shorttitle = {The {{US Geological Survey}} Ground Failure Product},
  author = {Allstadt, Kate E. and Thompson, Eric M. and Jibson, Randall W. and Wald, David J. and Hearne, Mike and Hunter, Edward J. and Fee, Jeremy and Schovanec, Heather and Slosky, Daniel and Haynie, Kirstie Lafon},
  year = 2022,
  journal = {Earthquake Spectra},
  volume = {38},
  number = {1},
  pages = {5--36},
  doi = {10.1177/87552930211032685},
  urldate = {2026-09-16},
  langid = {english}
}

@inproceedings{gerard_wildfirespreadts_2023,
  title = {{{WildfireSpreadTS}}: {{A}} Dataset of Multi-Modal Time Series for Wildfire Spread Prediction},
  shorttitle = {{{WildfireSpreadTS}}},
  booktitle = {Thirty-Seventh {{Conference}} on {{Neural Information Processing Systems Datasets}} and {{Benchmarks Track}}},
  author = {Gerard, Sebastian and Zhao, Yu and Sullivan, Josephine},
  year = 2023,
  month = nov,
  urldate = {2026-09-16},
  langid = {english}
}

@misc{gomez_tcbench_2026a,
  title = {{{TCBench}}: {{A Benchmark}} for {{Tropical Cyclone Track}} and {{Intensity Forecasting}} at the {{Global Scale}}},
  shorttitle = {{{TCBench}}},
  author = {Gomez, Milton and McGraw, Marie and S, Saranya Ganesh and Tam, Frederick Iat-Hin and Azizi, Ilia and Darmon, Samuel and Feldmann, Monika and Bourdin, Stella and {Poulain--Auz{\'e}au}, Louis and Camargo, Suzana J. and Lin, Jonathan and Chavas, Dan and Lee, Chia-Ying and Gupta, Ritwik and Jenney, Andrea and Beucler, Tom},
  year = 2026,
  month = jun,
  number = {arXiv:2601.23268},
  eprint = {2601.23268},
  primaryclass = {cs.CE},
  publisher = {arXiv},
  doi = {10.48550/arXiv.2601.23268},
  urldate = {2026-09-16},
  archiveprefix = {arXiv}
}

@inproceedings{huang_mlagentbench_2024,
  title = {{{MLAgentBench}}: {{Evaluating Language Agents}} on {{Machine Learning Experimentation}}},
  shorttitle = {{{MLAgentBench}}},
  booktitle = {Proceedings of the 41st {{International Conference}} on {{Machine Learning}}},
  author = {Huang, Qian and Vora, Jian and Liang, Percy and Leskovec, Jure},
  year = 2024,
  month = jul,
  pages = {20271--20309},
  publisher = {PMLR},
  issn = {2640-3498},
  urldate = {2026-09-16},
  langid = {english}
}

@article{khan_sharing_2019,
  title = {Sharing Interoperable Workflow Provenance: {{A}} Review of Best Practices and Their Practical Application in {{CWLProv}}},
  shorttitle = {Sharing Interoperable Workflow Provenance},
  author = {Khan, Farah Zaib and {Soiland-Reyes}, Stian and Sinnott, Richard O and Lonie, Andrew and Goble, Carole and Crusoe, Michael R},
  year = 2019,
  month = nov,
  journal = {GigaScience},
  volume = {8},
  number = {11},
  pages = {giz095},
  issn = {2047-217X},
  doi = {10.1093/gigascience/giz095},
  urldate = {2026-09-16}
}

@article{leijnse_modeling_2021,
  title = {Modeling Compound Flooding in Coastal Systems Using a Computationally Efficient Reduced-Physics Solver: {{Including}} Fluvial, Pluvial, Tidal, Wind- and Wave-Driven Processes},
  shorttitle = {Modeling Compound Flooding in Coastal Systems Using a Computationally Efficient Reduced-Physics Solver},
  author = {Leijnse, Tim and {van Ormondt}, Maarten and Nederhoff, Kees and {van Dongeren}, Ap},
  year = 2021,
  month = jan,
  journal = {Coastal Engineering},
  volume = {163},
  pages = {103796},
  issn = {0378-3839},
  doi = {10.1016/j.coastaleng.2020.103796},
  urldate = {2026-09-16}
}

@misc{mcgovern_extreme_2026,
  title = {Extreme {{Weather Bench}}: {{A}} Framework and Benchmark for Evaluation of High-Impact Weather},
  shorttitle = {Extreme {{Weather Bench}}},
  author = {McGovern, Amy and Mandelbaum, Taylor and Rothenberg, Daniel and Loveday, Nicholas and Potvin, Corey and Flora, Montgomery and Magnusson, Linus and Gilleland, Eric and Allen, John},
  year = 2026,
  month = may,
  number = {arXiv:2605.01126},
  eprint = {2605.01126},
  primaryclass = {cs.LG},
  publisher = {arXiv},
  doi = {10.48550/arXiv.2605.01126},
  urldate = {2026-09-16},
  archiveprefix = {arXiv}
}

@misc{shinn_reflexion_2023,
  title = {Reflexion: {{Language Agents}} with {{Verbal Reinforcement Learning}}},
  shorttitle = {Reflexion},
  author = {Shinn, Noah and Cassano, Federico and Berman, Edward and Gopinath, Ashwin and Narasimhan, Karthik and Yao, Shunyu},
  year = 2023,
  month = oct,
  number = {arXiv:2303.11366},
  eprint = {2303.11366},
  primaryclass = {cs.AI},
  publisher = {arXiv},
  doi = {10.48550/arXiv.2303.11366},
  urldate = {2026-09-16},
  archiveprefix = {arXiv}
}

@misc{gemma_gemma_2026,
  title = {Gemma 4 {{Technical Report}}},
  author = {Gemma, Team and Abd, Sherif El and Aggarwal, Vaibhav and Algayres, Robin and Andreev, Alek and Bachem, Olivier and Ballantyne, Ian and Brick, Cormac and C{\u a}rbune, Victor and Casbon, Michelle and Chaturvedi, Mayank and Chawla, Aditya and Cotruta, Victor and Coucke, Alice and Culliton, Phil and Dadashi, Robert and Dixon, Lucas and Elhawaty, Mohamed and Evci, Utku and Farabet, Cl{\'e}ment and Ferret, Johan and Galgani, Filippo and Girgin, Sertan and Grill, Jean-Bastien and Grootendorst, Maarten and Guo, Jiaxian and Hardin, Cassidy and He, Yanzhang and Hernandez, Steven M. and Homburger, Omri and Hussenot, L{\'e}onard and Ji, Juyeong and Joulin, Armand and Kamath, Aishwarya and Kassraie, Parnian and Lacombe, Olivier and Lahoti, Preethi and Liu, Ga{\"e}l and Martins, Gus and Martins, Luciano and Matejovicova, Tatiana and Merhej, Ramona and Momchev, Nikola and Mondal, Sneha and Mullins, Ryan and Panyam, Sindhu Raghuram and Pathak, Shreya and Perrin, Sarah and Pinto, Andr{\'e} Susano and Pot, Etienne and Pouget, Ang{\'e}line and Ram{\'e}, Alexandre and Ramos, Sabela and Reid, Douglas and Rim, David and Rivi{\`e}re, Morgane and Roth, Karsten and Rouillard, Louis and Sanseviero, Omar and Sessa, Pier Giuseppe and Settle, Shane and Sinopalnikov, Danila and Smoot, Sara and Stanczyk, Piotr and Steiner, Andreas and Stewart, Lawrence and Tolstikhin, Ilya and Tschannen, Michael and Tsitsulin, Anton and Vieillard, Nino and Wu, Renjie and Xu, Pingmei and Yang, Haichuan and Yvinec, Edouard and Zhang, Biao and Zhang, Li and Zou, Joe and Aagnes, Nicolas and Abdelhamed, Abdelrahman and Adamek, Jakub and Agrawal, Shivani and Agrawal, Shubham and Alabdulmohsin, Ibrahim and Alayrac, Jean Baptiste and Alon, Uri and Amarnath, Chandramouli and Anand, Ankesh and Anastasiou, Chrysovalantis and Ariafar, Setareh and Aubet, Fran{\c c}ois-Xavier and Axiotis, Kyriakos and Barbero, Federico and Barral, Joelle and Bendebury, Alexei and Bergmann, Urs and Bileschi, Stanley and Black, Kat and Blondel, Mathieu and Borgeaud, Sebastian and Bra{\v z}inskas, Arthur and Burnell, Ryan and {Busa-Fekete}, Robert and Cai, Mu and Calandriello, Daniele and Cameron, Glenn and Caucheteux, Charlotte and Chaabouni, Rahma and Chadha, Garima and Chan, Jetha and Chen, Blake Jianhang and Chen, Jesse and Chen, Lin and Chen, Xu and Cheng, Derek and Chien, Tzu-hsiang and Chinaev, Nikolai and Chou, Yi and Chu, Zhaohui and Coleman, Benjamin and Consul, Pooja and {Conway-Rahman}, Sam and Crowell, Scott and Cutler, Dylan and Dani, Vivek and Daruki, Samira and Das, Anil and Deutsch, Daniel and Dikkala, Nishanth and Ding, Li and Ding, Qiuhan and Dodhia, Shenil and Donhauser, Konstantin and Doshi, Tulsee and Dragan, Anca and Druinsky, Alex and Dua, Sahil and Egyed, Zoltan and Eisenbud, Danielle and Eppens, Daniel and Fan, Cindy and Fatemi, Bahare and Fathullah, Yassir and Feinberg, Vlad and Ferev, Milen and Flennerhag, Sebastian and Fujimoto, Takumi and Oliveira, Jo{\~a}o Gabriel and {Galatzer-Levy}, Isaac and Gante, Jo{\~a}o and Geisler, Simon and Ghosal, Soham and Girgis, Antonious M. and von Glehn, Tamara and Go, Alec and Gokhale, Alhaad and Grills, Alex and Gu, Yiming and Gupta, Mayank and Gupta, Pramod and Guruganesh, Guru and Hadsell, Raia and Harkous, Hamza and Harlalka, Jitendra and Hassabis, Demis and Hauth, Anja and Heyward, Joe and Hosseini, Arian and Hsia, Chih-Yang and Hsu, I.-Hung and Huang, Xiaopeng and Huang, Yangsibo and Hui, Kevin and Hutter, Adrian and I, Te and Iliopoulos, Fotis and Jain, Advait and Jawahar, Ganesh and Ji, Ziwei and Jin, Qilin and Johnson, Melvin and Joshi, Kandarp and Kandoor, Arun and Kang, Wang-Cheng and Kavukcuoglu, Koray and Kazemi, Mehran and Kenealy, Kathleen and Khalifa, Amr and Kirk, Phoebe and Korotkov, Ivan and Kothawade, Suraj and Kovalev, Vitaly and Kovelamudi, Neel and Kraft, Adam and Kumar, Ravin and Kumar, Vivek and Kuppam, Harish and Lannin, Justin and Lee, Chen-Yu and Lee, Seungji and Lepikhin, Dmitry and Levkovitch, Alon and Li, Dongdong and Li, Qiujia and Li{\'e}vin, Valentin and Lin, Ethan and Lin, Ziqian and Liu, Casper and Liu, Tianlin and Liu, Tianqi and Liu, Xin and Lobov, Ivan and Lunayach, Mayank and Ma, Min and Madan, Gagan and Maksai, Andrii and Malmi, Eric and Matuszak, Michal and McDuff, Daniel and Menghani, Gaurav and Miku{\l}a, Maciej and Mirylenka, Daniil and Misiunas, Karolis and Misra, Vedant and Mitran, Andreea and Mohamed, Kareem and Mukha, Maksim and Noland, Eric and O'Donnell, James and O'Donoghue, Brendan and Olszewska, Kate and Orlando, Bernett and Pan, Wanqiong and Panigrahy, Rina and Parekh, Unnati and {Perez-Nieves}, Nicolas and Park, Chunjong and Paskie, Eric and Peng, Liqian and Petrini, Bryce and Petrov, Slav and Pfeiffer, Jonas and Piot, Bilal and Plomecka, Martyna and Poder, Siim and Ponce, Octavio and Pramanik, Arijit and Racz, David and Rajan, Anish and Ramanovich, Michelle and Rao, Anand and Ritter, Marvin and Rodrigues, Vitor and Rosen, Evan and Rybi{\'n}ski, Miko{\l}aj and Sachdeva, Noveen and Sander, Micha{\"e}l E. and Sathyanarayana, Rohit and Savla, Sagar and Schmidgall, Samuel and Schuster, Tal and Scrivener, George and Seguin, Benoit and Sellergren, Andrew and Severyn, Aliaksei and Shafran, Izhak and Shah, Dhruv and Shahriari, Bobak and Shangguan, Yuan and Shenoy, Ashish and Shenoy, Pradeep and Shivanna, Rakesh and Sho, Pauline and Spangher, Lucas and Stokowiec, Wojciech and Strother, Tim and Su, Yao and Sun, Yinghao and Sundararajan, Mukund and Tacchetti, Andrea and Taege, Mor Hazan and Tafti, Pouya and Tarbouriech, Jean and Tekur, Chetan and Thakoor, Shantanu and Thapa, Rahul and Traverse, Madeleine and Treven, Lenart and Tu, Tao and Tung, Chien Te and {\"U}nl{\"u}, {\c C}a{\u g}lar and Veli{\v c}kovi{\'c}, Petar and Venkat, Malini Pooni and Venkatesh, Sagar Gubbi and Venkiteswaran, Vidya and Visin, Francesco and Vitvitskyi, Alex and Vodrahalli, Kiran and Wang, Weiyi and Wang, Xin and Warkentin, Tris and Wassenberg, Jan and Wieting, John and Wu, Cindy and Xiao, Lechao and Xu, Hao and Xu, Yuhui and Xue, Fuzhao and Yadav, Arun and Yan, Jun and Yang, Antoine and Yang, Lin and Yang, Ming-Hsuan and Ying, Ziyu and Yoo, Jae Hyeon and Zadimoghaddam, Morteza and Zafar, Sajjad and Zhang, Fred and Zhang, Jiageng and Zhang, Jianyi and Zhang, Xiaofan and Zhao, Chao and Zhou, David and Zou, Chen},
  year = 2026,
  month = jul,
  number = {arXiv:2607.02770},
  eprint = {2607.02770},
  primaryclass = {cs.CL},
  publisher = {arXiv},
  doi = {10.48550/arXiv.2607.02770},
  urldate = {2026-09-16},
  archiveprefix = {arXiv}
}

@inproceedings{trivedi_appworld_2024,
  title = {{{AppWorld}}: {{A Controllable World}} of {{Apps}} and {{People}} for {{Benchmarking Interactive Coding Agents}}},
  shorttitle = {{{AppWorld}}},
  booktitle = {Proceedings of the 62nd {{Annual Meeting}} of the {{Association}} for {{Computational Linguistics}} ({{Volume}} 1: {{Long Papers}})},
  author = {Trivedi, Harsh and Khot, Tushar and Hartmann, Mareike and Manku, Ruskin and Dong, Vinty and Li, Edward and Gupta, Shashank and Sabharwal, Ashish and Balasubramanian, Niranjan},
  editor = {Ku, Lun-Wei and Martins, Andre and Srikumar, Vivek},
  year = 2024,
  month = aug,
  pages = {16022--16076},
  publisher = {Association for Computational Linguistics},
  address = {Bangkok, Thailand},
  doi = {10.18653/v1/2024.acl-long.850},
  urldate = {2026-09-16}
}

@misc{valmeekam_planbench_2023,
  title = {{{PlanBench}}: {{An Extensible Benchmark}} for {{Evaluating Large Language Models}} on {{Planning}} and {{Reasoning}} about {{Change}}},
  shorttitle = {{{PlanBench}}},
  author = {Valmeekam, Karthik and Marquez, Matthew and Olmo, Alberto and Sreedharan, Sarath and Kambhampati, Subbarao},
  year = 2023,
  month = nov,
  number = {arXiv:2206.10498},
  eprint = {2206.10498},
  primaryclass = {cs.CL},
  publisher = {arXiv},
  doi = {10.48550/arXiv.2206.10498},
  urldate = {2026-09-16},
  archiveprefix = {arXiv}
}

@article{woollam_seisbench_2022,
  title = {{{SeisBench}}---{{A Toolbox}} for {{Machine Learning}} in {{Seismology}}},
  author = {Woollam, Jack and M{\"u}nchmeyer, Jannes and Tilmann, Frederik and Rietbrock, Andreas and Lange, Dietrich and Bornstein, Thomas and Diehl, Tobias and Giunchi, Carlo and Haslinger, Florian and Jozinovi{\'c}, Dario and Michelini, Alberto and Saul, Joachim and Soto, Hugo},
  year = 2022,
  month = mar,
  journal = {Seismological Research Letters},
  volume = {93},
  number = {3},
  pages = {1695--1709},
  issn = {0895-0695},
  doi = {10.1785/0220210324},
  urldate = {2026-09-16}
}

@article{xu_floodcastbench_2025,
  title = {{{FloodCastBench}}: {{A Large-Scale Dataset}} and {{Foundation Models}} for {{Flood Modeling}} and {{Forecasting}}},
  shorttitle = {{{FloodCastBench}}},
  author = {Xu, Qingsong and Shi, Yilei and Zhao, Jie and Zhu, Xiao Xiang},
  year = 2025,
  month = mar,
  journal = {Scientific Data},
  volume = {12},
  number = {1},
  pages = {431},
  publisher = {Nature Publishing Group},
  issn = {2052-4463},
  doi = {10.1038/s41597-025-04725-2},
  urldate = {2026-09-16},
  copyright = {2025 The Author(s)},
  langid = {english}
}

@misc{king_post-wildfire_2025,
  title = {Post-Wildfire Debris-Flow Hazard Assessment ({{PWFDF}}) Collection},
  author = {King, Jonathan M. and Kostelnik, Jaime and Kean, Jason W. and Barnhart, Katherine R. and Graber, Andrew P. and Rengers, Francis K. and Thomas, Matthew A.},
  year = {2025},
  publisher = {{U.S. Geological Survey}},
  institution = {{Geologic Hazards Science Center}},
  type = {Data Release},
  doi = {10.5066/P14EWYME},
  url = {https://doi.org/10.5066/P14EWYME},
  urldate = {2026-09-16},
  note = {CC0 1.0 Universal}
}

@misc{noaa_cpc_spi_2026,
  title = {Standardized Precipitation Index},
  author = {{NOAA Climate Prediction Center}},
  year = {2026},
  publisher = {{National Oceanic and Atmospheric Administration}},
  url = {https://www.cpc.ncep.noaa.gov/products/Drought/Monitoring/spi.shtml},
  urldate = {2026-09-16},
  note = {Operational drought-monitoring product based on the CPC Unified Precipitation Analysis, 1950--present}
}

@misc{nasa_global_landslide_nowcast_2025,
  title = {Global Landslide Nowcast from {{LHASA L4}} 1 Day 1 km $\times$ 1 km, Version 2.0.0},
  shorttitle = {Global Landslide Nowcast from {{LHASA}}, Version 2.0.0},
  author = {{NASA Goddard Earth Sciences Data and Information Services Center}},
  year = {2025},
  publisher = {{NASA Goddard Earth Sciences Data and Information Services Center}},
  type = {Dataset},
  doi = {10.5067/8VKQDQFFOTS3},
  url = {https://doi.org/10.5067/8VKQDQFFOTS3},
  urldate = {2026-09-16},
  note = {Dataset identifier: Global\_Landslide\_Nowcast; Level 4; daily temporal resolution; 1 km spatial resolution}
}

@misc{noaa_ncei_storm_events_1996,
  title = {{{Storm Events Database}}},
  author = {{NOAA National Centers for Environmental Information}},
  year = {1996},
  publisher = {{National Oceanic and Atmospheric Administration}},
  type = {Ongoing Dataset},
  url = {https://www.ncei.noaa.gov/access/storm-events-database/},
  urldate = {2026-09-16},
  note = {Version 1.0; NCEI dataset identifier 3910\_03; continuously updated}
}

@inproceedings{trirat_automlagent_2025,
  title = {{{AutoML-Agent}}: {{A Multi-Agent LLM Framework}} for {{Full-Pipeline AutoML}}},
  shorttitle = {{{AutoML-Agent}}},
  booktitle = {Proceedings of the 42nd {{International Conference}} on {{Machine Learning}}},
  author = {Trirat, Patara and Jeong, Wonyong and Hwang, Sung Ju},
  year = 2025,
  month = oct,
  pages = {60099--60146},
  publisher = {PMLR},
  issn = {2640-3498},
  urldate = {2026-09-21},
  langid = {english}
}

@inproceedings{wang_planandsolve_2023,
  title = {Plan-and-{{Solve Prompting}}: {{Improving Zero-Shot Chain-of-Thought Reasoning}} by {{Large Language Models}}},
  shorttitle = {Plan-and-{{Solve Prompting}}},
  booktitle = {Proceedings of the 61st {{Annual Meeting}} of the {{Association}} for {{Computational Linguistics}} ({{Volume}} 1: {{Long Papers}})},
  author = {Wang, Lei and Xu, Wanyu and Lan, Yihuai and Hu, Zhiqiang and Lan, Yunshi and Lee, Roy Ka-Wei and Lim, Ee-Peng},
  editor = {Rogers, Anna and {Boyd-Graber}, Jordan and Okazaki, Naoaki},
  year = 2023,
  month = jul,
  pages = {2609--2634},
  publisher = {Association for Computational Linguistics},
  address = {Toronto, Canada},
  doi = {10.18653/v1/2023.acl-long.147},
  urldate = {2026-09-21}
}

@inproceedings{yao_tree_2023,
  title = {Tree of {{Thoughts}}: {{Deliberate Problem Solving}} with {{Large Language Models}}},
  shorttitle = {Tree of {{Thoughts}}},
  booktitle = {Advances in {{Neural Information Processing Systems}}},
  author = {Yao, Shunyu and Yu, Dian and Zhao, Jeffrey and Shafran, Izhak and Griffiths, Tom and Cao, Yuan and Narasimhan, Karthik},
  year = 2023,
  volume = {36},
  pages = {11809--11822},
  publisher = {Curran Associates, Inc.},
  doi = {10.52202/075280-0517},
  urldate = {2026-09-21}
}

@inproceedings{chan_mlebench_2024,
  title = {{{MLE-bench}}: {{Evaluating Machine Learning Agents}} on {{Machine Learning Engineering}}},
  shorttitle = {{{MLE-bench}}},
  booktitle = {The {{Thirteenth International Conference}} on {{Learning Representations}}},
  author = {Chan, Jun Shern and Chowdhury, Neil and Jaffe, Oliver and Aung, James and Sherburn, Dane and Mays, Evan and Starace, Giulio and Liu, Kevin and Maksin, Leon and Patwardhan, Tejal and Madry, Aleksander and Weng, Lilian},
  year = 2024,
  month = oct,
  urldate = {2026-09-25},
  langid = {english}
}

@inproceedings{lewis_retrievalaugmented_2020,
  title = {Retrieval-{{Augmented Generation}} for {{Knowledge-Intensive NLP Tasks}}},
  booktitle = {Advances in {{Neural Information Processing Systems}}},
  author = {Lewis, Patrick and Perez, Ethan and Piktus, Aleksandra and Petroni, Fabio and Karpukhin, Vladimir and Goyal, Naman and K{\"u}ttler, Heinrich and Lewis, Mike and Yih, Wen-tau and Rockt{\"a}schel, Tim and Riedel, Sebastian and Kiela, Douwe},
  year = 2020,
  volume = {33},
  pages = {9459--9474},
  publisher = {Curran Associates, Inc.},
  urldate = {2026-09-25}
}

@inproceedings{wang_selfconsistency_2023,
  title = {Self-{{Consistency Improves Chain}} of {{Thought Reasoning}} in {{Language Models}}},
  booktitle = {The {{Eleventh International Conference}} on {{Learning Representations}}},
  author = {Wang, Xuezhi and Wei, Jason and Schuurmans, Dale and Le, Quoc V. and Chi, Ed H. and Narang, Sharan and Chowdhery, Aakanksha and Zhou, Denny},
  year = 2023,
  month = sep,
  urldate = {2026-09-25},
  langid = {english}
}
\bibliographystyle{iclr2027_conference}

\clearpage
\appendix
\clearpage
\appendix
\section*{Appendix}

% \xq{There are only tables in the appendix; let's turn some simple and small experiment results table into plots, especially those wrapped tables}
\counterwithin{table}{section}
\counterwithin{figure}{section}

\section{HWB Scope and Benchmark Construction}
\label{app:hwb-scope}

This appendix provides the benchmark composition, implementation details, baseline adaptations, evaluation protocol, and supplementary results supporting the main paper. All reported results use the same sealed 141-instance inventory and the DCA evaluator.

\subsection{Benchmark Scope}

HWB contains 141 sealed instances spanning seven single-hazard domains and four multi-hazard interaction classes. The final inventory contains 90 single-hazard and 51 multi-hazard instances. At the initial state, 45 instances admit one registered route and 96 expose multiple candidate routes.

% \begin{wraptable}[18]{r}{0.55\linewidth}
% \centering
% \vspace{-0.9em}
% \caption{\textbf{Sealed HWB inventory.} The final column gives the number of scored instances.}
% \vspace{-0.5em}
% \label{tab}
% \setlength{\tabcolsep}{2.8pt}
% \renewcommand{\arraystretch}{1.08}
% \begin{tabular*}{\linewidth}
% {@{\extracolsep{\fill}}llr@{}}
% \toprule
% Track & Scientific target & $N$ \\
% \midrule
% WF-3   & Active-fire spread                & 13 \\
% FL-2   & Dynamic flood inundation          & 12 \\
% L2     & Rainfall-triggered landslide      & 14 \\
% E1--E3 & Seismic phase, source, and motion & 14 \\
% TC-TRK & Tropical-cyclone trajectory       &  6 \\
% DR-OUT & Drought outlook                   & 14 \\
% HW-MED & Medium-range heatwave             & 17 \\
% MH-1   & Post-fire debris-flow cascade     & 16 \\
% MH-2   & Earthquake-triggered landslide    & 12 \\
% MH-3   & Coastal compound flooding         & 11 \\
% MH-4   & Multi-peril impact                & 12 \\
% \midrule
% Total  &                                   & 141 \\
% \bottomrule
% \end{tabular*}
% % \vspace{-0.8cm}
% \end{wraptable}

\begin{wrapfigure}[16]{r}{0.45\linewidth}
\vspace{-0.8cm}
\centering
\includegraphics[
  width=\linewidth,
  keepaspectratio
]{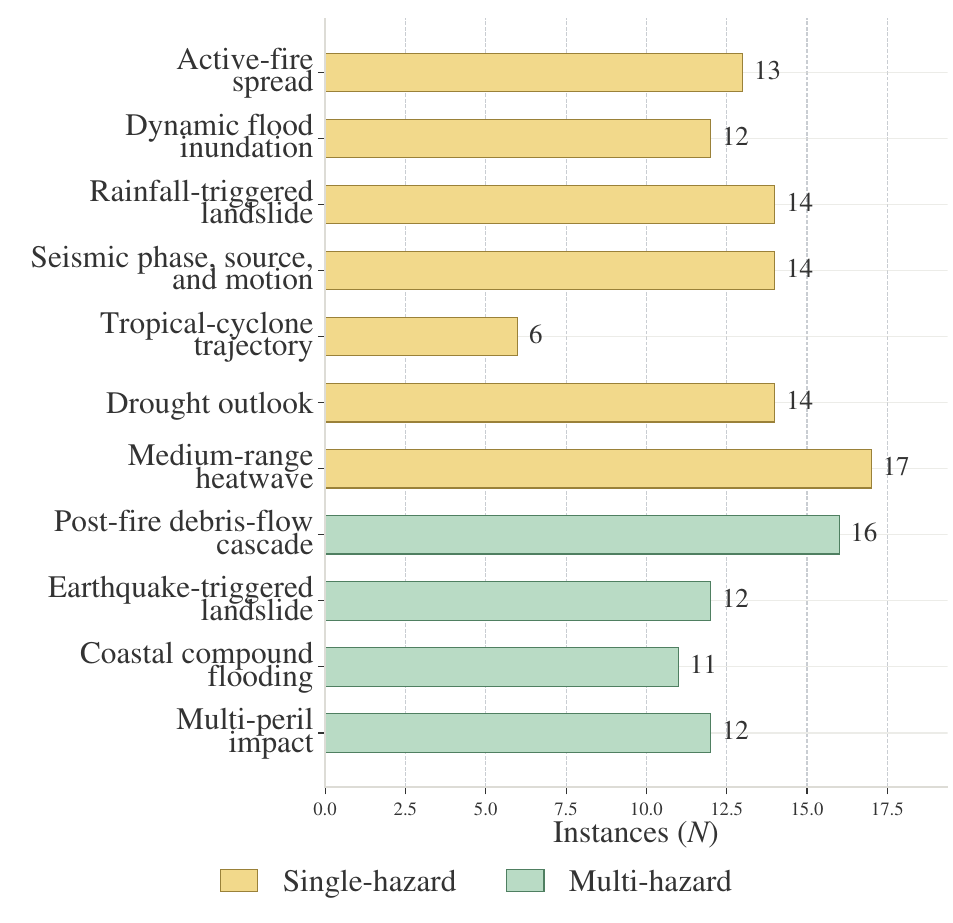}
\vspace{-0.6cm}
\caption{HWB inventory.
Instance counts across the 7 single- and 4 multi-hazard tracks.}
\label{fig:appendix-hwb-inventory}
\vspace{-0.8cm}
\end{wrapfigure}
The construction infrastructure begins from a global atlas of 3,062 registered geospatial scenarios, of which 542 were materialized for benchmark generation. The atlas is a construction resource: headline results are computed exclusively on the sealed 141-instance inventory. The selected instances preserve the diversity of their scientific sources while binding every task to an explicit output contract, candidate route set, capability state, and task-native evaluator.

The eleven source families provide observed events, forcing variables, operational products, and reference outputs for benchmark construction. They support wildfire, flood, landslide, earthquake, tropical-cyclone, drought, heatwave, and multi-hazard tasks across geographically diverse settings.

\begin{table*}[t]
\centering
\caption{\textbf{Scientific data sources used in HWB construction.}
Each source supplies observed events, forcing variables, reference products, or evaluator targets; the corresponding publications and official records are cited in Section~4.}
\label{tab:appendix-data-sources}
\setlength{\tabcolsep}{5pt}
\renewcommand{\arraystretch}{1.12}
\begin{tabular*}{\textwidth}
{@{\extracolsep{\fill}}
>{\raggedright\arraybackslash}p{0.25\textwidth}
>{\raggedright\arraybackslash}p{0.73\textwidth}@{}}
\toprule
Hazard task & Scientific source and role \\
\midrule
Wildfire spread
& \textbf{WildfireSpreadTS.} Time-indexed fire observations for evaluating the spatial and temporal progression of active-fire spread. \\

Flood inundation
& \textbf{FloodCastBench.} Hydrological forcing fields and reference flood extents for dynamic inundation analysis. \\

Rainfall-triggered landslide
& \textbf{NASA LHASA v2.} Global landslide nowcasts and reference occurrences for rainfall-triggered susceptibility analysis. \\

Earthquake analysis
& \textbf{SeisBench.} Waveforms and reference labels for phase picking, event association, and seismic-source estimation. \\

Tropical-cyclone tracking
& \textbf{TCBench.} Cyclone trajectories and lead-time-specific references for evaluating forecast track error. \\

Drought outlook
& \textbf{NOAA CPC drought products.} Standardized precipitation indicators and operational references for drought assessment. \\

Heatwave forecasting
& \textbf{ExtremeWeatherBench.} High-impact weather events and reference targets for medium-range forecast verification. \\

Post-fire debris flow
& \textbf{USGS Post-Wildfire Debris-Flow Hazard Assessment Collection.} Burn severity, rainfall, likelihood, and volume for cascade analysis. \\

Earthquake-triggered ground failure
& \textbf{USGS Ground Failure products.} Landslide and liquefaction estimates for ground-failure and coupled-impact assessment. \\

Coastal compound flooding
& \textbf{SFINCS coastal simulations.} Inundation targets under interacting oceanic, pluvial, and fluvial forcing conditions. \\

Multi-peril impact
& \textbf{NOAA Storm Events Database and HWB damage cases.} Event-level records for property-damage assessment across interacting hazards. \\
\bottomrule
\end{tabular*}
\end{table*}

\subsection{Hazard and Route Strata}

% \begin{wraptable}[5]{r}{0.50\linewidth}
% \centering
% \vspace{-4em}
% \caption{Joint benchmark strata.}
% \vspace{-0.5em}
% \label{tab:appendix-joint-strata}
% \setlength{\tabcolsep}{3.0pt}
% \renewcommand{\arraystretch}{1.10}
% \begin{tabular*}{\linewidth}
% {@{\extracolsep{\fill}}lrrr@{}}
% \toprule
% & One-route & Multi-route & Total \\
% \midrule
% Single hazard & 29 & 61 & 90 \\
% Multi hazard  & 16 & 35 & 51 \\
% \midrule
% Total         & 45 & 96 & 141 \\
% \bottomrule
% \end{tabular*}
% \vspace{-0.8em}
% \end{wraptable}
\begin{wrapfigure}[11]{r}{0.45\linewidth}
\vspace{-0.7cm}
\centering
\includegraphics[
  width=\linewidth,
  keepaspectratio
]{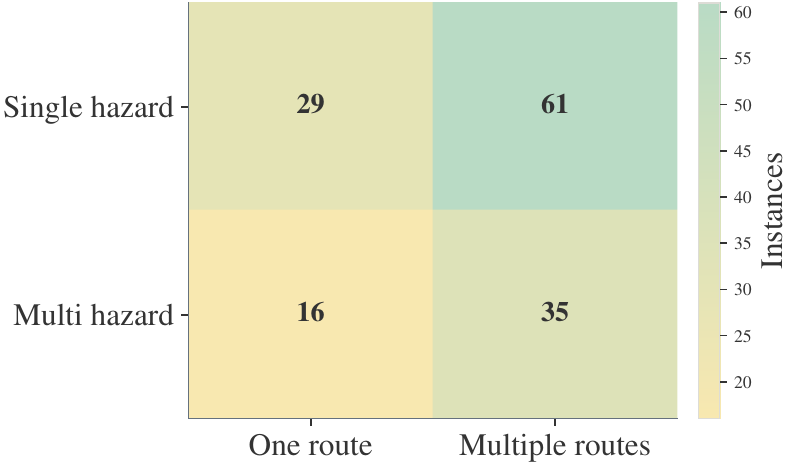}
\vspace{-0.6cm}
\caption{Joint benchmark strata.
Instance counts by hazard and route  composition.}
\label{fig:appendix-joint-strata}
\vspace{-0.8em}
\end{wrapfigure}
The hazard-composition and route-multiplicity strata are independent. Figure~\ref{fig:appendix-joint-strata} reports their joint distribution, allowing performance to be examined both by scientific composition and by the amount of route choice presented to the agent. Multiple-route cases occur in both single- and multi-hazard settings.

Each instance specifies a scientific target, initial state, information cutoff, candidate route set, capability state, output contract, and task-native evaluator. The 96 multiple-route instances expose more than one registered candidate at the initial state. The evaluator checks the committed route against the benchmark-side criterion bound to the corresponding scenario.

\subsection{Construction and Sealing}

Benchmark construction proceeds through four stages. First, scientific datasets and operational products are mapped to task definitions with explicit output variables, units, spatial and temporal support, and evaluation criteria. Second, candidate route families are instantiated from Route Cards and executable capabilities. Third, every candidate instance undergoes schema, provenance, temporal-cutoff, and replay checks. Finally, the selected inventory, route registry, capability versions, evaluator contracts, and reference artifacts are frozen together.

Every sealed instance has at least one construction-time workflow that can produce the required artifact in a clean environment. This workflow establishes solvability during construction; it is not shown to the evaluated agent and does not prescribe its execution trajectory.

\section{Route Cards, Capabilities, and Agent Operations}
\label{app:route-representation}

\subsection{Route Card Representation}

A Route Card records the scientific and executable information required to determine whether a route is applicable and realizable. Its principal fields are summarized in Table~\ref{tab:route-card-fields}. Full HazardWeaver resolves the frozen HKC-derived conditions associated with these Route Cards at the initial state and revisits their status when permitted evidence changes. The HKC-off condition in RQ2 removes this evidence-linked applicability layer while retaining the remainder of the evaluation pipeline.

\begin{table}[t]
\centering
\caption{{Principal Route Card fields.}}
\label{tab:route-card-fields}
\setlength{\tabcolsep}{4pt}
\renewcommand{\arraystretch}{1.08}
\begin{tabular*}{\linewidth}{@{\extracolsep{\fill}}lp{0.68\linewidth}@{}}
\toprule
Field group & Contents \\
\midrule
Task identity
& Hazard domain, task specification, requested output, and route family. \\

Scientific conditions
& Assumptions, required observations, validity ranges, calibration regime, scale, and uncertainty treatment. \\

Capability requirements
& Required datasets, models, formulas, transformations, and compatibility or dependency constraints. \\

Output contract
& Variable definition, units, format, spatial and temporal support, resolution, and artifact schema. \\

Evaluation
& Task-native metric, tolerance, and acceptance threshold. \\

Provenance
& Scientific source, capability version, supporting evidence span, and benchmark-construction record. \\
\bottomrule
\end{tabular*}
\end{table}

\subsection{Typed Capability Representation}

HCG contains 82 registered capability specifications corresponding to 79 distinct executable capabilities. Capabilities include scientific datasets, empirical equations, numerical models, learned models, simulators, transformations, and coupling operators. Each input and output port records the scientific variable, unit, representation, spatial and temporal support, resolution, uncertainty metadata, and provenance requirements. Representation-preserving transfers may connect capabilities directly. Operations that change information, including interpolation, aggregation, thresholding, learned downscaling, and unit conversion, remain explicit capabilities. Their effects therefore remain visible in the executed workflow and its provenance record.

\subsection{Permitted Agent Operations}

Table~\ref{tab:recovery-actions} lists the grounded operations used by HWA. Every operation either inspects the current state, executes a registered capability, or produces an auditable state transition.

\begin{table*}[t]
\centering
\caption{Grounded tools available to HWA. Operations are instantiated only when permitted by the task and remaining execution state.}
\label{tab:recovery-actions}
\setlength{\tabcolsep}{5pt}
\renewcommand{\arraystretch}{1.10}
\begin{tabular*}{\textwidth}{@{\extracolsep{\fill}}p{0.22\textwidth}p{0.74\textwidth}@{}}
\toprule
Operation & Purpose and state effect \\
\midrule
Inspect state       & Reads evidence, artifacts, and capability status; no state change. \\
Inspect route       & Reads route conditions, capabilities, and contract; no state change. \\
Execute capability  & Runs a capability; records artifact and provenance, or the failure. \\
Request information & Obtains a permitted unresolved observation; adds it as evidence. \\
Repair workflow     & Inserts a transformation or subpath; keeps the route-level target. \\
Change route        & Commits to another route from the recomputed eligible set. \\
Submit artifact     & Returns the terminal artifact and invokes the sealed evaluator. \\
Submit abstention   & Abstains after failed recovery; invokes the abstention criterion. \\
\bottomrule
\end{tabular*}
\end{table*}

\section{Evaluation and Reporting Protocol}
\label{app:evaluation-protocol}

\subsection{Decision-Constrained Accuracy}

DCA follows Eq.~\ref{eq:dca} in the main paper. For an artifact submission, a pass requires task-native artifact success together with benchmark-side route validity, provenance validity, typed and causal execution validity, and the applicable resource contract. An abstention is credited only under the registered recovery criterion in Eq.~\ref{eq:abstention}. Missing, incomplete, or unevaluated submissions remain in the denominator and count as failures.

For reporting, $E_q$ denotes the artifact-validity factor in Eq.~\ref{eq:dca}, while $V_q$ aggregates the corresponding trajectory-validity factors. Table~\ref{tab:rq3-main} reports their marginal pass rates; headline DCA requires them jointly with benchmark-side route validity. Consequently, $E_q$ or $V_q$ may exceed DCA when the remaining criterion is not satisfied.

\subsection{Task-Native Evaluators}

HWB preserves the scientific evaluator appropriate to each output rather than forcing heterogeneous artifacts into a common numerical error. Table~\ref{tab:track-evaluators} summarizes the evaluator families. Exact thresholds, output fields, and units are stored with the sealed instance contracts.

\begin{table*}[t]
\centering
\caption{Task-native evaluator families. DCA applies the artifact criteria below jointly with route and execution validity.}
\label{tab:track-evaluators}
\setlength{\tabcolsep}{5pt}
\renewcommand{\arraystretch}{1.12}
\begin{tabular*}{\textwidth}
{@{\extracolsep{\fill}}
>{\raggedright\arraybackslash}p{0.3\textwidth}
>{\raggedright\arraybackslash}p{0.7\textwidth}@{}}
\toprule
Hazard track & Scientific outputs and evaluation \\
\midrule

Active-fire spread
& Fire probability, occupancy, or perimeter, evaluated using average precision, intersection over union, and registered field checks. \\

Dynamic flood inundation
& Inundation depth and extent, evaluated using depth RMSE over the scenario-specific spatial support. \\

Rainfall-triggered landslide
& Landslide occurrence probability, evaluated using area under the precision--recall curve. \\

Seismic phase, source, and motion
& Phase picks, event associations, source parameters, and ground motion, evaluated using pick F1, association score, and coverage. \\

Tropical-cyclone trajectory
& Forecast storm trajectories, evaluated using lead-time-specific track error in kilometers. \\

Drought outlook
& Drought categories or probabilities, evaluated using the registered operational outlook-skill criterion. \\

Medium-range heatwave
& Heatwave probabilities or intensities, evaluated against the maximum accepted forecast error. \\

Post-fire debris-flow cascade
& Debris-flow likelihood, volume, or affected area, evaluated using output-specific volume error, Brier score, precision--recall. \\

Earthquake-triggered landslide
& Triggered-landslide or service-impact outputs, evaluated through scientific replay under the registered coupling model. \\

Coastal compound flooding
& Compound-flood depth and extent, evaluated using depth RMSE under the combined forcing conditions. \\

Multi-peril impact
& Multi-peril loss or substrate coverage, evaluated using registered log-scale error and coverage criteria. \\

\bottomrule
\end{tabular*}
\end{table*}

\subsection{Statistical Reporting}

Instance-level proportions use Wilson 95\% confidence intervals. Macro DCA is the unweighted mean of the 11 track-level DCA rates; its 95\% interval is obtained by resampling the 11 tracks with replacement for 5,000 bootstrap replicates using seed 42. Comparisons between HazardWeaver and individual baselines use an exact paired McNemar test over the same 141 sealed instances.

\section{Implementation Details}
\label{app:implementation-details}

\subsection{Language-Model Backbones}

\begin{wraptable}[10]{r}{0.7\linewidth}
\vspace{-0.6cm}
\centering
\caption{Policy backbones. Model identifiers used with Full HazardWeaver; all non-policy components remain fixed.}
\vspace{-0.5em}
\label{tab:backbone-ids}
\setlength{\tabcolsep}{2.5pt}
\renewcommand{\arraystretch}{1.08}
\begin{tabular*}{\linewidth}
{@{\extracolsep{\fill}}
>{\raggedright\arraybackslash}p{0.34\linewidth}
>{\raggedright\arraybackslash}p{0.59\linewidth}@{}}
\toprule
Backbone & Model identifier \\
\midrule
OLMo-3.1-32B
& allenai/Olmo-3.1-32B-Instruct \\

Mixtral-8x22B
& mistralai/Mixtral-8x22B-Instruct-v0.1 \\

Gemma-4-31B-it
& google/gemma-4-31B-it \\

Llama-3.3-70B
& meta-llama/Llama-3.3-70B-Instruct \\

DeepSeek-V4.1-Flash
& deepseek-v4.1-flash \\
\bottomrule
\end{tabular*}
\vspace{-0.8em}
\end{wraptable}
Experiments use seed 11, temperature 0.2, the frozen HKC registry, the matched execution protocol, and DCA rescoring. All compared methods operate on the same sealed inventory, candidate route catalog, scientific execution interface, evaluator contracts, and per-instance failure policy. A run that does not return a valid scored submission counts as a failure.

\subsection{Execution Records}

Each run stores the selected route, route commitments, capability invocations, generated artifacts, provenance edges, state transitions, and terminal submission. Numerical, spatial, and probabilistic outputs are generated by registered scientific capabilities; the policy model controls route selection and execution flow.

\section{Baseline Adaptations}
\label{app:baseline-details}

The source systems target different execution environments and task types. We therefore adapt their published decision principles to the common HWB route-selection interface. Every method receives the same task description, candidate Route Cards, registered capability identifiers, public output requirements, and scientific execution interface. After route selection, the chosen route is executed and evaluated through the same sealed replay and DCA stack used for HazardWeaver.

The adaptations preserve the characteristic control strategy of each method while avoiding method-specific scientific tools, private reference artifacts, or evaluator feedback during route selection. Table~\ref{tab:baseline-adaptations} summarizes the resulting implementations.

\begin{table*}[t]
\centering
\caption{{Baseline adaptations used in RQ1.} ``Rounds'' refers to route-selection or reflection rounds before the terminal commitment.}
\label{tab:baseline-adaptations}
\setlength{\tabcolsep}{5pt}
\renewcommand{\arraystretch}{1.10}
\begin{tabular*}{\textwidth}
{@{\extracolsep{\fill}}p{0.15\textwidth}p{0.8\textwidth}@{}}
\toprule
Method & Adapted principle and HWB realization \\
\midrule
AIDE
& Iterative candidate refinement. Refines a proposed scientific route for up to three rounds using the task and Route Cards, then commits directly. \\

DS-Agent
& Case-based retrieval. Retrieves the three most similar training-side route cases and uses them to rank the registered candidates. \\

R\&D-Agent
& Hypothesis-driven ordering. Forms a task hypothesis, orders the candidate scientific routes against it, and commits to the highest-ranked route. \\

MLE-STAR
& Metric-directed program search. Infers the required scientific metric and output program before ranking candidate routes against those requirements. \\

AutoML-Agent
& Pipeline and schema selection. Infers the task input--output schema and ranks routes by their ability to realize a compatible scientific pipeline. \\

DisasterBench-ToT
& Deliberate search over disaster tools. Treats registered capabilities as typed disaster tools and performs bounded tree search over candidate route choices. \\

ReAct
& Interleaved reasoning and action. Alternates route inspection and comparison for up to three turns before issuing a final commitment. \\

Reflexion
& Verbal reflection from execution outcomes. Revises its route choice for up to three rounds using a qualitative failure class from the preceding execution. \\

Self-Consistency
& Sampling and majority agreement. Produces five independent route rankings and commits to the candidate receiving the largest number of first-place votes. \\

Plan-and-Solve
& Explicit planning before execution. Produces an ordered scientific plan and executes its first committed route under the shared interface. \\
\bottomrule
\end{tabular*}
\end{table*}

These adaptations separate route-selection strategy from scientific execution. Their outputs remain fully comparable because the selected capabilities are evaluated on the same instances with the same artifact and route contracts.

\section{Supplementary Experimental Results}
\label{app:additional-results}

\subsection{Exact RQ2 Ablation Results}

\begin{wraptable}[9]{r}{0.49\linewidth}
\centering
\vspace{-3em}
\caption{Scientific-constraint ablations. DCA on the 96 route-choice instances; entries give count and percentage.}
\vspace{-0.5em}
\label{tab:appendix-rq2}
\setlength{\tabcolsep}{2.5pt}
\renewcommand{\arraystretch}{1.10}
\begin{tabular*}{\linewidth}
{@{\extracolsep{\fill}}lccc@{}}
\toprule
Backbone & HKC off & HCG untyped & Full \\
\midrule
OLMo
& 38 (39.6) & 39 (40.6) & \textbf{49 (51.0)} \\
Mixtral
& 34 (35.4) & 32 (33.3) & \textbf{65 (67.7)} \\
Gemma
& 50 (52.1) & 55 (57.3) & \textbf{76 (79.2)} \\
DeepSeek
& 68 (70.8) & 71 (74.0) & \textbf{86 (89.6)} \\
Llama
& 44 (45.8) & 46 (47.9) & \textbf{90 (93.8)} \\
\bottomrule
\end{tabular*}
\vspace{-0.8em}
\end{wraptable}

Table~\ref{tab:appendix-rq2} gives the counts underlying Figure~\ref{fig:rq2-constraints}. The 96 instances are the multiple-route stratum of the sealed 141-instance benchmark. Removing either evidence-linked applicability conditions or typed capability composition reduces DCA for every backbone.

The route-update ablation in Table~\ref{tab:rq2-route-update} uses 40 applicable instances drawn from the sealed pool. Across five backbones, updated eligibility passes 83 of 200 evaluated backbone--cell combinations, compared with 32 of 200 under fixed initial eligibility. This pooled value is descriptive; the per-backbone comparisons remain the primary result.

\subsection{Paired Significance Tests}

% \begin{wraptable}[14]{r}{0.55\linewidth}
% \centering
% \vspace{-1.cm}
% \caption{\textbf{Paired significance tests.} Discordant DCA for HazardWeaver and each baseline.}
% \vspace{-0.5em}
% \label{tab:appendix-mcnemar}
% \setlength{\tabcolsep}{2.8pt}
% \renewcommand{\arraystretch}{1.07}
% \begin{tabular*}{\linewidth}
% {@{\extracolsep{\fill}}lrrr@{}}
% \toprule
% Baseline & HW-only & BL-only & Exact $p$ \\
% \midrule
% Plan-and-Solve & 75 & 7 & $1.37{\times}10^{-13}$ \\
% Self-Consistency & 80 & 4 & $2.76{\times}10^{-16}$ \\
% DisasterBench-ToT & 80 & 4 & $2.76{\times}10^{-16}$ \\
% MLE-STAR          & 80 & 1 & $4.45{\times}10^{-18}$ \\
% AutoML-Agent      & 93 & 5 & $1.52{\times}10^{-18}$ \\
% Reflexion         & 89 & 8 & $4.56{\times}10^{-16}$ \\
% AIDE              & 94 & 0 & $8.62{\times}10^{-22}$ \\
% DS-Agent          & 82 & 3 & $2.67{\times}10^{-17}$ \\
% R\&D-Agent        & 82 & 4 & $1.01{\times}10^{-16}$ \\
% ReAct             & 84 & 0 & $1.35{\times}10^{-19}$ \\
% \bottomrule
% \end{tabular*}
% \vspace{-0.8em}
% \end{wraptable}
\begin{wrapfigure}[16]{r}{0.50\linewidth}
\vspace{-.65cm}
\centering
\includegraphics[
  width=\linewidth,
  keepaspectratio
]{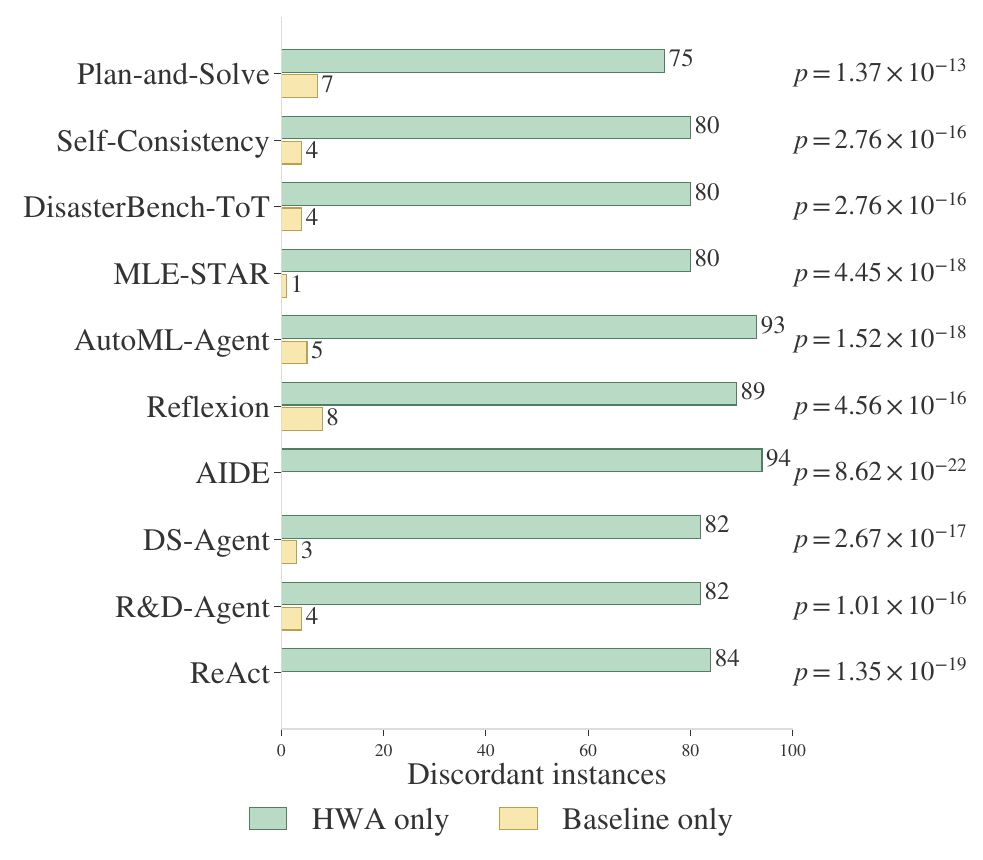}
\vspace{-0.75cm}
\caption{Paired significance tests.
Discordant DCA counts between HazardWeaver and each baseline; annotations give McNemar $p$-values.}
\label{fig:appendix-mcnemar}
\vspace{-0.8em}
\end{wrapfigure}
The test is paired at the instance level: for each baseline, HWA-only counts cases solved exclusively by HazardWeaver, whereas BL-only counts cases solved exclusively by that baseline. This construction controls for variation in task difficulty because both systems are evaluated on exactly the same 141 instances. The imbalance is substantial for every comparison. Even against Plan-and-Solve, the strongest baseline by aggregate DCA, HazardWeaver uniquely solves 75 instances while losing only seven in the opposite direction. The pattern becomes still more pronounced for the shared-tool controllers; for example, HazardWeaver records 84 unique successes against ReAct with no ReAct-only success. All exact McNemar tests remain significant by a wide margin.

These paired results complement the aggregate comparison in Table~\ref{tab:main-results}. The advantage is not produced by averaging over a small number of unusually favorable tracks or by one isolated family of tasks. Instead, the discordant outcomes consistently favor HazardWeaver across the sealed inventory and across baseline designs based on planning, reflection, retrieval, iterative refinement, and sampled agreement. This provides instance-level evidence that explicit scientific route eligibility improves the reliability of the resulting decisions beyond the particular control strategy used for comparison.

\subsection{Joint Hazard and Route Strata}

% \begin{wraptable}[11]{r}{0.4\linewidth}
% \centering
% \vspace{-1em}
% \caption{\textbf{Joint-stratum DCA counts.} S/M denote single-/multi-hazard; 1/R denote one-/multiple-route.}
% \vspace{-0.5em}
% \label{tab:appendix-joint-results}
% \setlength{\tabcolsep}{3.0pt}
% \renewcommand{\arraystretch}{1.10}
% \begin{tabular*}{\linewidth}
% {@{\extracolsep{\fill}}lrrr@{}}
% \toprule
% Stratum & HWA & Plan & AIDE \\
% \midrule
% S $\times$ 1 ($N=29$) & 24 & 13 & 7 \\
% S $\times$ R ($N=61$) & 58 & 24 & 13 \\
% M $\times$ 1 ($N=16$) & 12 & 13 & 6 \\
% M $\times$ R ($N=35$) & 32 & 8 & 6 \\
% \bottomrule
% \end{tabular*}
% \vspace{-0.8em}
% \end{wraptable}

\begin{wrapfigure}[14]{r}{0.60\linewidth}
\vspace{-0.8em}
\centering
\includegraphics[
  width=\linewidth,
  keepaspectratio
]{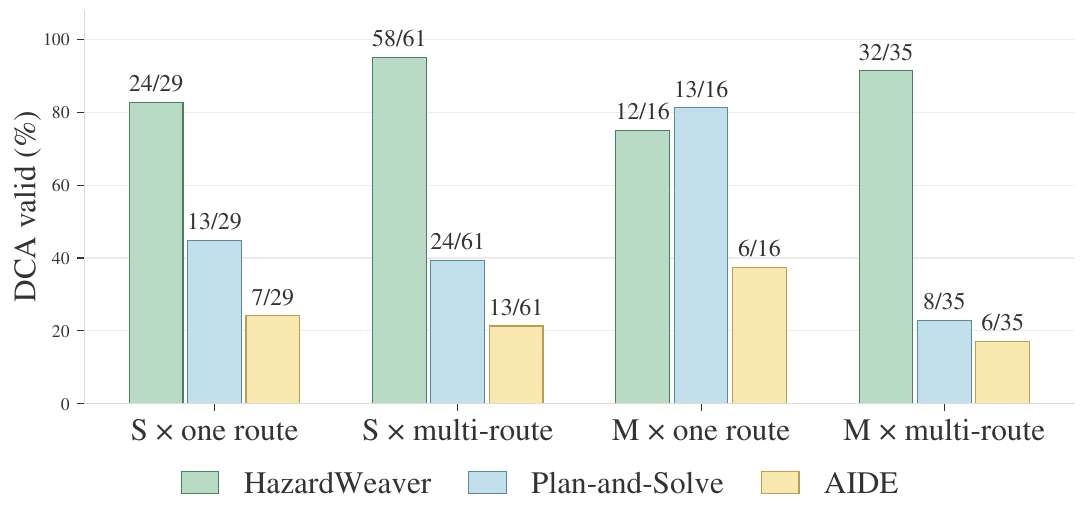}
\vspace{-0.75cm}
\caption{Joint-stratum performance.
DCA across hazard composition and route multiplicity; annotations report successful instances over stratum size.}
\label{fig:appendix-joint-results}
\vspace{-0.8em}
\end{wrapfigure}
Figure~\ref{fig:appendix-joint-results} separates hazard composition from route multiplicity instead of treating either property as a proxy for the other. HazardWeaver performs strongly in both route-rich strata, solving 58 of 61 single-hazard and 32 of 35 multi-hazard instances with multiple candidate routes. Its advantage is therefore preserved when route choice is combined with a composed scientific analysis. Plan-and-Solve is competitive on the smaller multi-hazard, one-route subset, where it solves 13 of 16 instances compared with 12 for HazardWeaver. This local result also shows that the overall comparison is not uniformly favorable by construction. The largest separation appears once the task requires choosing among several routes. Plan-and-Solve falls from 13 of 16 on multi-hazard, one-route tasks to eight of 35 on the corresponding multiple-route stratum, whereas HazardWeaver solves 32 of those 35 instances. AIDE shows a similar difference between the route regimes. Together with the controlled ablations in RQ2, this stratification supports that HazardWeaver's principal advantage lies in organizing scientific route choice while retaining strong execution across both isolated and composed hazard analyses.

\subsection{Track-Level Cross-Backbone Results}

\begin{figure*}[t]
    \centering
    \includegraphics[width=\textwidth]{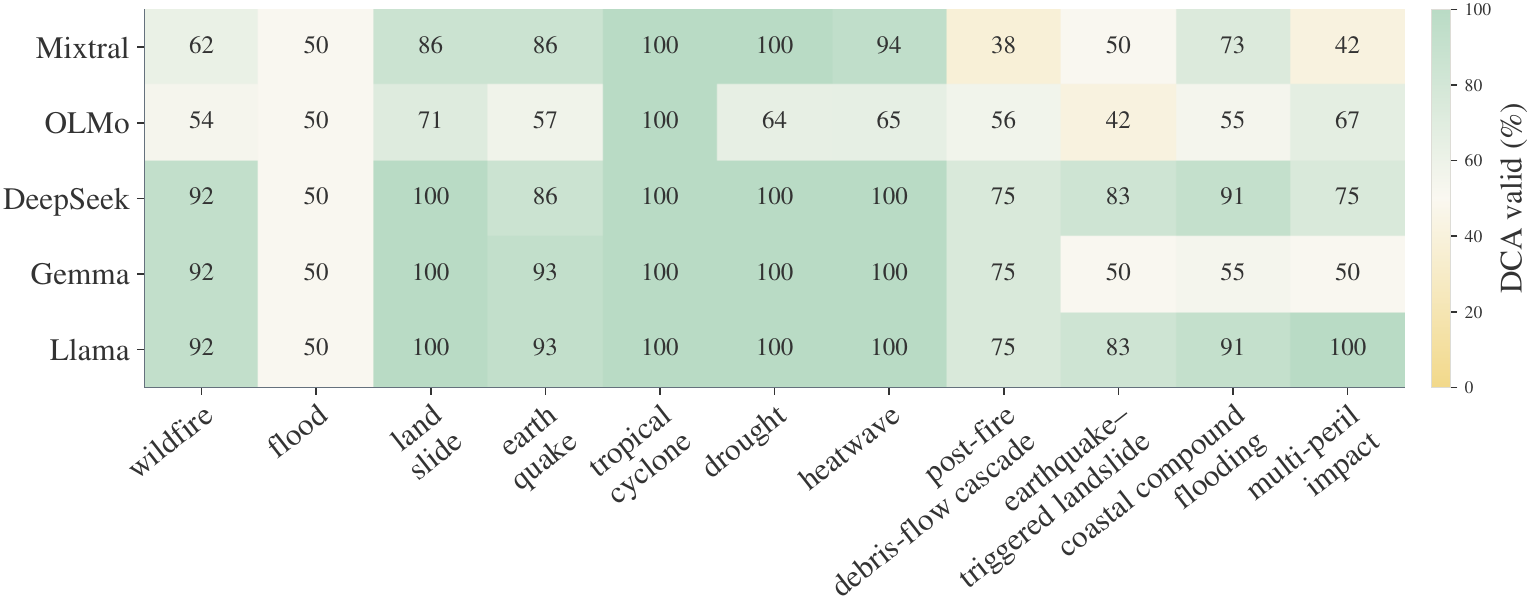}
    \vspace{-0.75cm}
    \caption{Track-level DCA for Full HazardWeaver. Each entry reports track-level DCA (\%).}
    \label{fig:appendix-track-results}
    \vspace{-0.4cm}
\end{figure*}

% \begin{table*}[t]
% \centering
% \caption{{}
% \label{tab:appendix-track-results}
% \setlength{\tabcolsep}{5pt}
% \renewcommand{\arraystretch}{1.08}
% \begin{tabular*}{\textwidth}{@{\extracolsep{\fill}}lccccc@{}}
% \toprule
% Track & OLMo & Mixtral & Gemma & DeepSeek & Llama \\
% \midrule
% WF-3   & 7/13  & 8/13  & 12/13 & 12/13 & 12/13 \\
% FL-2   & 6/12  & 6/12  & 6/12  & 6/12  & 6/12 \\
% L2     & 10/14 & 12/14 & 14/14 & 14/14 & 14/14 \\
% E1--E3 & 8/14  & 12/14 & 13/14 & 12/14 & 13/14 \\
% TC-TRK & 6/6   & 6/6   & 6/6   & 6/6   & 6/6 \\
% DR-OUT & 9/14  & 14/14 & 14/14 & 14/14 & 14/14 \\
% HW-MED & 11/17 & 16/17 & 17/17 & 17/17 & 17/17 \\
% MH-1   & 9/16  & 6/16  & 12/16 & 12/16 & 12/16 \\
% MH-2   & 5/12  & 6/12  & 6/12  & 10/12 & 10/12 \\
% MH-3   & 6/11  & 8/11  & 6/11  & 10/11 & 10/11 \\
% MH-4   & 8/12  & 5/12  & 6/12  & 9/12  & 12/12 \\
% \midrule
% Overall & 85/141 & 99/141 & 112/141 & 122/141 & 126/141 \\
% \bottomrule
% \end{tabular*}
% \end{table*}

Performance is distributed across all 11 tracks. The tropical-cyclone trajectory track is solved by every backbone, while the dynamic flood inundation track remains a shared bottleneck. The multi-hazard tracks produce greater cross-backbone separation, consistent with the broader profiles shown in Figure~\ref{fig:rq3-tracks}.

\section{Benchmark Sealing and Quality Control}
\label{app:sealing-qa}

\subsection{Evaluator Validation}

Before sealing, the evaluator was exercised on 160 controlled fixtures spanning eight decision classes: valid alternative execution, scientifically inapplicable route, capability-incompatible route, justified abstention, unjustified abstention, post-cutoff information use, prediction without capability provenance, and coupling-insensitive multi-hazard output. All 160 fixtures produced their expected accept or reject decision.

\subsection{Information Separation}

\begin{wraptable}[8]{r}{0.4\linewidth}
\centering
\vspace{-2.cm}
\caption{\textbf{Information separation.} Availability for execution and scoring.}
\vspace{-1em}
\label{tab:information-separation}
\setlength{\tabcolsep}{3.0pt}
\renewcommand{\arraystretch}{1.10}
\begin{tabular*}{\linewidth}
{@{\extracolsep{\fill}}lcc@{}}
\toprule
Information & Agent & Eval. \\
\midrule
Task and initial state       & \checkmark & \checkmark \\
Candidate routes             & \checkmark & \checkmark \\
Capability interface         & \checkmark & \checkmark \\
Public output contract       & \checkmark & \checkmark \\
Reference artifact           &            & \checkmark \\
Sealed route criterion       &            & \checkmark \\
Construction workflow        &            & \checkmark \\
\bottomrule
\end{tabular*}
\vspace{-0.8em}
\end{wraptable}

Reference artifacts, evaluator thresholds beyond the public task specification, construction-time workflows, and benchmark-side route judgments are held outside the solver context. Agents receive the task, initial state, candidate route descriptions, permitted capabilities, and public output requirements. Evaluation is performed only after terminal submission. All benchmark-side route judgments are frozen before any evaluated policy run; evaluation never derives or revises its route criterion from the tested system's output.

\subsection{Route-Registry Review}

We conducted a stratified human review of 86 route--condition pairs drawn from 44 benchmark cells across all 11 tracks. All 27 sampled witness routes were judged scientifically applicable. Candidates rejected by the sealed contract were subsequently adjudicated against the requested output, scientific semantics, and capability requirements; this review found no route that was both contract-valid and omitted from the registry. We use this review as quality control for the sealing procedure rather than as a claim that the catalog exhausts every route that future scientific work may establish.

\section{Representative Execution Traces}
\label{app:execution-traces}

\subsection{Route Revision After an Execution-State Change}

In a representative seismic instance, HWA initially enumerates the available seismic routes, commits to the capability satisfying the phase-picking contract, and begins execution. An intermediate state change invalidates the current realization. HWA updates the capability state, reinstantiates a compatible realization, recommits under the revised state, and completes the required artifact. The final route and phase-picking output both satisfy the sealed evaluator.

This trace illustrates the distinction between retaining a scientific route family and repairing its executable realization: the scientific target remains appropriate, while the capability path must be reconstructed from the updated state.

\subsection{Failure Under an Incomplete Cascade}

A representative post-fire debris-flow cascade failure occurs when the policy selects a debris-flow volume route for a task requiring the complete burn--rainfall--debris-flow cascade. The selected capability produces a scientifically meaningful intermediate quantity but does not satisfy the terminal route and output contract. The evaluator therefore rejects the submission. This case shows that successful execution of an individual scientific model does not by itself establish a decision-valid multi-stage analysis.

\section{Reproducibility and Release Contents}
\label{app:reproducibility}

The release is organized around the sealed inventory so that every reported aggregate can be reconstructed from per-instance records. Table~\ref{tab:release-contents} summarizes the principal artifacts.

\begin{table*}[t]
\centering
\caption{{Reproducibility artifacts accompanying HWB.}}
\label{tab:release-contents}
\setlength{\tabcolsep}{5pt}
\renewcommand{\arraystretch}{1.10}
\begin{tabular*}{\textwidth}
{@{\extracolsep{\fill}}p{0.24\textwidth}p{0.66\textwidth}@{}}
\toprule
Artifact & Contents \\
\midrule
Sealed inventory
& The 141 instance identifiers, track assignments, task specifications, initial states, route strata, and evaluator bindings. \\

Route Cards
& Scientific conditions, required observations, capability signatures, output contracts, uncertainty fields, and evidence provenance. \\

Capability registry
& Executable entry points, versioned dependencies, typed input and output ports, transformations, and validation metadata. \\

Evaluator contracts
& Task-native metrics, tolerances, route criteria, cutoff rules, provenance requirements, and abstention conditions. \\

Run configurations
& Model identifiers, stochastic settings, execution protocol, and method-specific selector configuration. \\

Per-instance outputs
& Selected routes, state transitions, capability invocations, generated artifacts, provenance records, and terminal evaluation reports. \\

Aggregation scripts
& DCA, marginal validity, macro DCA, Wilson intervals, bootstrap intervals, and paired significance calculations. \\

Figure and table inputs
& Machine-readable values used to generate every experimental figure and table in the paper. \\
\bottomrule
\end{tabular*}
\end{table*}

A complete reproduction first materializes the frozen task and capability versions, executes the selected system configuration on all 141 instances, applies the sealed evaluator to each terminal record, and then aggregates the resulting per-instance decisions. The released table inputs provide an additional direct check that the manuscript values match the sealed evaluation outputs.

\section{Limitations}

HazardWeaver operates within a frozen scientific capability catalog and evidence index. HKC conditions reflect the literature corpus and Route Cards available at benchmark construction; new evidence, revised standards, or updated models may require recompilation and review. HWB evaluates artifacts and committed routes through frozen task-specific contracts, so reported validity denotes reproducible agreement with these documented criteria. Headline scoring covers routes registered when the benchmark is sealed; newly established routes require catalog extension and resealing. The 141 instances span eleven hazard tracks and diverse geographic settings without exhausting every event, region, sensor configuration, or operational constraint. Operational forecasting and emergency decision-making therefore require updated evidence, local calibration, and domain review.

\end{document}